\documentclass{article}
\usepackage[utf8]{inputenc}
\usepackage{main}
\usepackage{microtype}
\usepackage{graphicx}
\usepackage{times}
\usepackage{amsmath}
\usepackage{amssymb}
\usepackage{enumitem}
\usepackage{booktabs}
\usepackage{xcolor}     
\usepackage{natbib}
\definecolor{mydarkblue}{rgb}{0,0.08,0.45}
\usepackage[colorlinks=true,linkcolor=mydarkblue,citecolor=mydarkblue,filecolor=mydarkblue,urlcolor=mydarkblue]{hyperref}
\usepackage{fancyhdr}

\definecolor{gred}{RGB}{250, 210, 207}
\definecolor{coolblue1}{rgb}{0.91, 0.94, 0.98}
\definecolor{coolblue2}{rgb}{0.76, 0.85, 0.94}
\definecolor{coolblue3}{rgb}{0.54, 0.72, 0.87}
\definecolor{coolblue4}{rgb}{1, 1, 1}

\usepackage[table]{xcolor}  

\usepackage{xspace}
\usepackage{cleveref}
\usepackage[]{multicol, multirow}
\usepackage[most]{tcolorbox}
\usepackage{etoc}

\tcbuselibrary{listingsutf8}
\tcbuselibrary{listingsutf8,breakable}

\newtcolorbox[auto counter]{observation}[1][]{
  colback=black!5!white,
  colframe=black!70!white,
  fonttitle=\bfseries,
  title=Observation~\thetcbcounter,
  enhanced,
  boxrule=0.6pt,
  left=1mm,right=1mm,top=1mm,bottom=1mm,
  #1
}

\newtcolorbox[auto counter]{takeaway}[1][]{
  colback=teal!3!white,
  colframe=teal!55!black,
  fonttitle=\bfseries,
  title=Takeaway~\thetcbcounter,
  enhanced,
  boxrule=0.5pt,
  left=1mm,right=1mm,top=1mm,bottom=1mm,
  #1
}

\newtcolorbox[auto counter]{practicalguidance}[1][]{
  colback=cyan!3!white,
  colframe=cyan!60!black,
  fonttitle=\bfseries,
  title=Practical Guidance~\thetcbcounter,
  enhanced,
  boxrule=0.5pt,
  left=1mm,right=1mm,top=1mm,bottom=1mm,
  #1
}

\newtcolorbox[auto counter]{discussion}[1][]{
  colback=violet!4!white,
  colframe=violet!60!black,
  fonttitle=\bfseries,
  title=Discussion~\thetcbcounter,
  enhanced,
  boxrule=0.5pt,
  left=1mm,right=1mm,top=1mm,bottom=1mm,
  #1
}

\usepackage{caption}

\usepackage{color}
\usepackage{tabularx}
\usepackage{tabu}
\usepackage{booktabs}
\usepackage{colortbl}
\usepackage{multirow}

\usepackage{graphicx} 
\usepackage{float}
\usepackage{subfigure} 
\usepackage{amssymb}

\usepackage{hyperref}
\usepackage{tablefootnote}
\usepackage{xspace}

\usepackage{float}
\usepackage{amsmath}
\usepackage{bm}
\usepackage{algorithmicx}
\usepackage{algorithm}
\usepackage{algpseudocode}

\usepackage{arydshln}
\usepackage{amssymb}
\usepackage{pifont}
\usepackage{enumitem}

\newenvironment{itemize*}%
 {\leftmargini=10pt\begin{itemize}%
  \setlength{\itemsep}{0pt}%
  \setlength{\parskip}{0pt}%
  }%
 {\end{itemize}}
\newenvironment{enumerate*}%
 {\begin{enumerate}%
  \setlength{\itemsep}{0pt}%
  \setlength{\parskip}{0pt}}%
 {\end{enumerate}}

\usepackage[most]{tcolorbox}
\definecolor{myblue}{rgb}{0.18,0.45,0.73}

\usepackage[table]{xcolor}   
\definecolor{grpours}{RGB}{231,245,255}   
\definecolor{grpllm}{RGB}{240,240,240}    
\definecolor{grpother}{RGB}{237,250,233}  

\usepackage{pgfplots}
\pgfplotsset{compat=1.18}
\usepackage{xcolor}

\usepackage{booktabs}
\usepackage{siunitx}
\definecolor{customblue}{HTML}{286dc0}

\definecolor{customgreen}{HTML}{2ca02c}

\newtcolorbox{blueBox}[1][]{
  enhanced,
  colback=customblue!5!white,
  colframe=customblue,
  boxrule=0.8pt,
  arc=1mm,
  left=5pt,
  right=5pt,
  top=6pt,
  bottom=6pt,
  title={\centering #1},
  width=\linewidth
}

\newtcolorbox{analysisbox}[1][]{
    enhanced,
    colframe=customgreen,
    colback=customgreen!10!white,
    sharp corners,
    boxsep=0pt,
    left=5pt,
    right=5pt,
    top=6pt,
    bottom=6pt,
    boxrule=0pt,
    leftrule=4pt,
    width=\linewidth,
    #1
}

\usepackage{microtype}

\newcommand{\huggingface}{\raisebox{-1.5pt}{\includegraphics[height=1.05em]{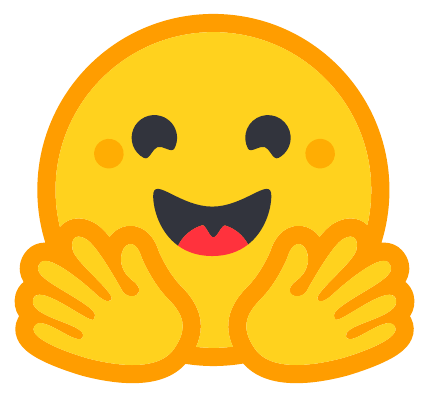}}\xspace}
\newcommand{\github}{\raisebox{-1.5pt}{\includegraphics[height=1.05em]{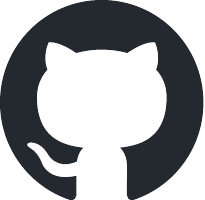}}\xspace}
\newcommand{\data}{RMR-75K}

\newcommand{\Yale}{\hspace{.1em}^{\textcolor{YaleBlue}{\boldsymbol{Y}}}}
\newcommand{\TCS}{\hspace{.1em}^{\textcolor{TCSC}{\boldsymbol{T}}}}
\newcommand{\NYU}{\hspace{.1em}^{\textcolor{NYUPurple}{\boldsymbol{N}}}}

\definecolor{YaleBlue}{RGB}{0, 53, 107}
\definecolor{NYUPurple}{RGB}{134, 1, 175}  
\definecolor{TCSC}{RGB}{1, 126, 199}

\begin{document}

\title{\textsc{RbtAct}: Rebuttal as Supervision for Actionable Review Feedback Generation}

\author{
\textbf{Sihong Wu}$\Yale$ \quad
\textbf{Yiling Ma}$\Yale$ \quad
\textbf{Yilun Zhao}$\Yale$\thanks{Correspondence to: Yilun Zhao (\texttt{yilun.zhao@yale.edu})} \quad
\textbf{Tiansheng Hu}$\NYU$ \quad
\textbf{Owen Jiang}$\Yale$ \\ [3pt]
\textbf{Manasi Patwardhan}$\TCS$ \quad
\textbf{Arman Cohan}$\Yale$ \\ [7pt]
 $\Yale$Yale University \quad $\NYU$New York University \quad $\TCS$TCS Research
}

\maketitle
\thispagestyle{fancy}
\fancyhead{}
\lhead{%
    \includegraphics[height=1.1cm]{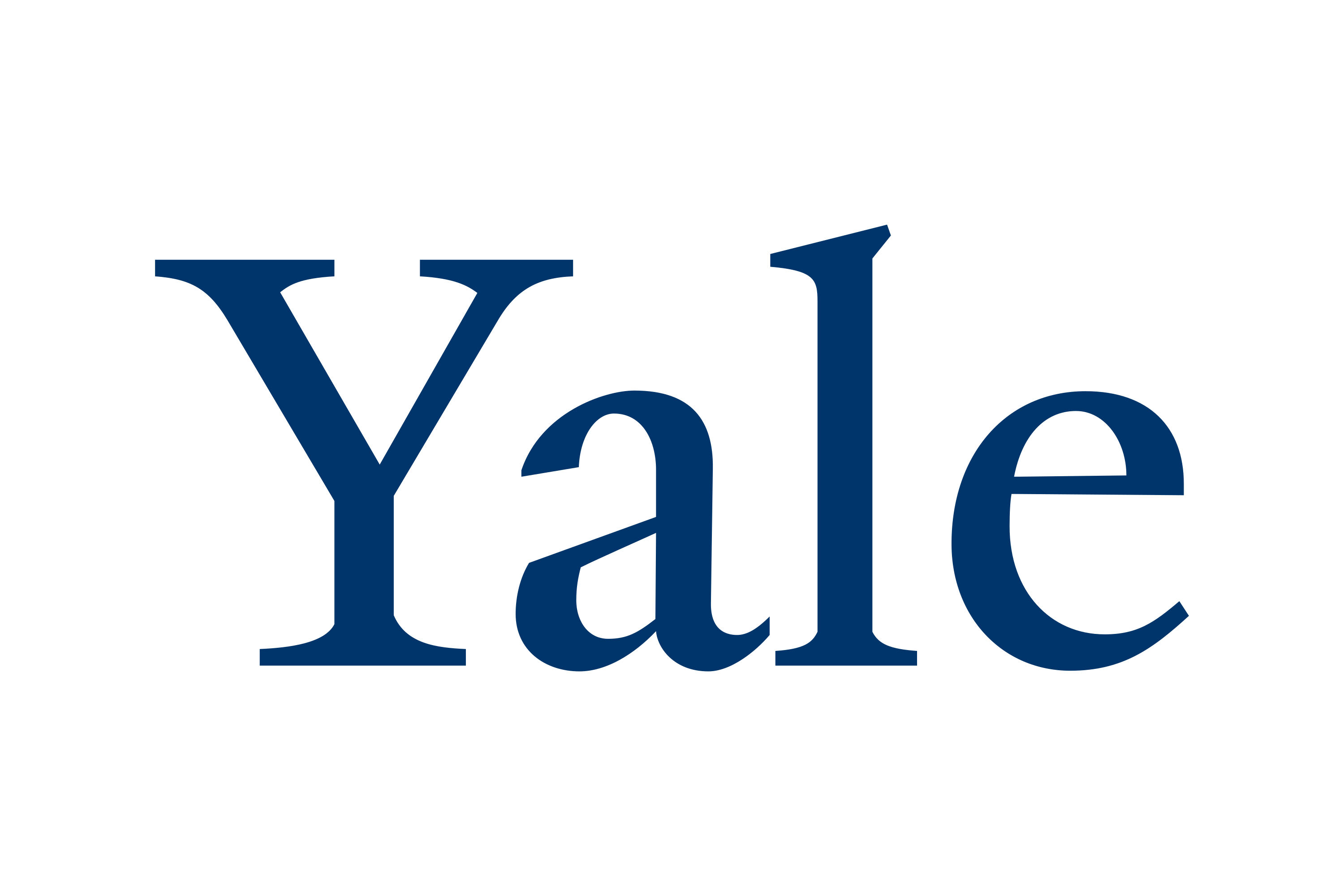}\hspace{0.2cm}%
    \raisebox{0.2cm}{\includegraphics[height=0.7cm]{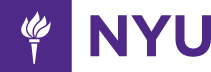}}\hspace{0.2cm}%
    \raisebox{0.05cm}{\includegraphics[height=0.8cm]{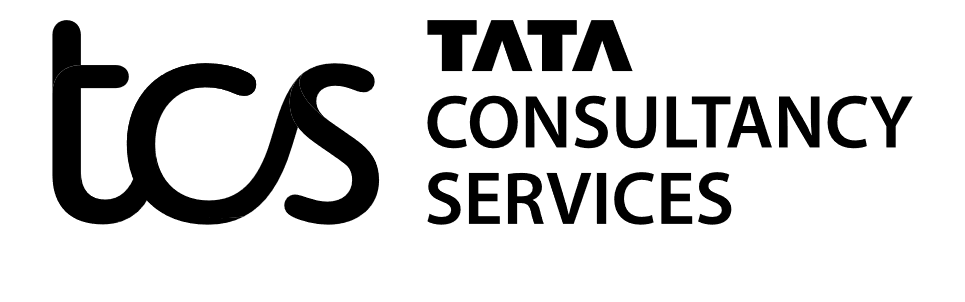}}%
}

\fancyfoot[C]{\thepage}
\renewcommand{\headrulewidth}{0pt}
\setlength{\headheight}{12pt}
\addtolength{\topmargin}{0pt}
\setlength{\headsep}{3mm}

\vspace{-1.0em}

\pagestyle{plain}

\begin{abstract}
Large language models (LLMs) are increasingly used across the scientific workflow, including to draft peer-review reports. However, many AI-generated reviews are superficial and insufficiently actionable, leaving authors without concrete, implementable guidance and motivating the gap this work addresses. We propose \textsc{RbtAct}, which targets actionable review feedback generation and places existing peer review rebuttal at the center of learning. Rebuttals show which reviewer comments led to concrete revisions or specific plans, and which were only defended. Building on this insight, we leverage rebuttal as implicit supervision to directly optimize a feedback generator for actionability. 
To support this objective, we propose a new task called \emph{perspective-conditioned segment-level review feedback generation}, in which the model is required to produce a single focused comment based on the complete paper and a specified perspective such as experiments and writing. We also build a large dataset named \data\ that maps review segments to the rebuttal segments that address them, with perspective labels and impact categories that order author uptake. We then train the Llama-3.1-8B-Instruct model with supervised fine-tuning on review segments followed by preference optimization using rebuttal derived pairs. Experiments with human experts and LLM-as-a-judge show consistent gains in actionability and specificity over strong baselines while maintaining grounding and relevance. 

\begin{center}
\begin{tabular}{cl@{\hspace{5em}}cl}
\huggingface & \href{https://huggingface.co/datasets/shwu22/RMR-75K}{\textbf{Data:} RMR-75K} &
\github & \href{https://github.com/formula12/RbtAct/}{\textbf{Code:} RbtAct}
\end{tabular}
\end{center}
\vspace{5pt}

\end{abstract}


\section{Introduction}
LLMs are increasingly used in scientific research, including assisting with scientific writing and peer reviews~\cite{zhao2025sciarena, zhu-etal-2025-deepreview}. Early work explored whether LLMs can help draft peer reviews or support reviewers through prompting methods \cite{gpt4sli,hosseini2023chatgptpeerreview, xu-etal-2025-llms-identify}. Subsequent research moved from prompting to fine-tuning methods \cite{reviewer2, cycleresearcher} with multi-agent coordination \citep{agentreview,tan2024peerreviewmultiturnlongcontext,deepreview}. Prior studies show that while LLMs can draft fluent reviews, they often miss specific issues, show shallow analysis, and produce generic phrasing, so their feedback does not reliably act as actionable guidance \cite{reviewergpt,blindspot}. 

At the same time, the peer review process itself contains a rich supervision signal that scrutinizes the scientific merits of a work and allows researchers to improve their research. This is often done through author rebuttals in the peer review process, where authors either commit to concrete changes or defer action on certain reviewer comments \cite{disapere}. We argue that rebuttals are an underutilized source of implicit human feedback for learning what kinds of comments actually trigger revision.\footnote{Reviews and rebuttals can be noisy or inconsistent, but we hypothesize that large-scale use provides sufficient signal to improve model training.} \
In this work, we propose \textsc{RbtAct}, utilizing rebuttals as an \emph{implicit preference signal} and using them to directly optimize review generation for \emph{actionability}. Concretely, we derive pairwise preferences from rebuttal outcomes and apply preference optimization so that the model favors comments that elicited author action \citep{dpo}. This turns rebuttal from an object of analysis \cite{re2} into a supervision source for training a model.

Typically, full conference reviews mix strengths, weaknesses, and questions across multiple aspects. Authors mainly respond to weaknesses and questions, which vary across perspectives such as experiments, novelty, writing~\cite{ghosal2022peer}. 
Treating a full review as one unit makes actionability difficult to evaluate because author reactions target only parts of the review. We therefore decompose reviews into key-point segments and study segment-level review feedback generation for a given perspective: given the full paper and a target perspective (e.g., \emph{Experiments}), the model produces one focused comment. This design narrows scope, promotes specificity, and enables precise supervision by aligning each review segment with the rebuttal segment that addresses it.

Prior work connects reviews to downstream author behavior in different ways. ARIES \cite{aries} links review comments to paper edits, enabling edit prediction from feedback but not aligning comments to rebuttal text. DISAPERE \cite{disapere} annotates discourse relations between review and rebuttal at the sentence level, but at a smaller scale and without perspective labels. Our setting complements both: we segment reviews into key points, align each point to the corresponding rebuttal span, and attach a perspective label and an impact category that captures the author’s reaction, such as a concrete revision performed, a planned revision, or a defense without changes. These impact categories make actionability concrete and induce pairwise preferences in which comments leading to revisions outrank those that yield plans or defenses. To formulate preference pairs, for each paper and perspective, we collect all review segments aligned to rebuttal spans; whenever two segments share the same paper and perspective but have different impact categories, we form a pair. Compared with previous datasets, our resource targets segment-level review feedback generation and provides rebuttal-anchored supervision that reflects what authors actually did or committed to do.


We first fine-tune Llama-3.1-8B-Instruct on perspective-conditioned review segments to establish a strong baseline. 
We then apply preference-optimization through rebuttal-derived DPO \cite{dpo} to optimize the model for actionability. This pipeline treats rebuttal as a natural reward model to serve as a signal for actionable review generation, targeting the gap identified by prior evaluations that LLM-generated reviews are often generic and not sufficiently tied to revision \citep{hosseini2023chatgptpeerreview,swift, sadallah2025goodbadconstructiveautomatically}.

Experiments show that our model produces review feedback that is more actionable and specific than competitive baselines under both human and LLM-as-a-judge evaluations. We evaluate on a test set built from ICLR 2025 and the improvements hold across multiple perspectives. Comparisons cover our \textsc{RbtAct}, an SFT-only variant, and larger prompted LLMs such as Llama-3.1-70B and GPT-5-chat. \textsc{RbtAct} achieves the highest actionability in both studies: 3.46 out of 5 in human evaluation and 3.38 out of 5 in LLM-as-a-judge evaluation, maintaining parity on groundedness and relevance, with fine-grained analyses across seven perspectives and pairwise win rates indicating consistent gains.

We summarize our contributions as follows:
\begin{itemize} [leftmargin=*]
\itemsep0em 
\item We present the framework \textsc{RbtAct} that first utilizes author rebuttals as implicit supervision and applies preference optimization to directly optimize a feedback generator for actionability.

\item We release a large-scale dataset named \data, which contains 75,542 examples. Each example consists of (i) a review segment, (ii) an associated perspective on the review, (iii) an author response to the review, and (iv) an annotated impact category indicating the review’s actionability.

\item We propose an effective training pipeline that yields consistent gains in actionability and other aspects over competitive baselines under human and LLM-as-a-judge evaluations.

\end{itemize}

\begin{figure*}[t]
    \vspace*{-2em} 
    \centering
	\includegraphics[width=1.0\linewidth]{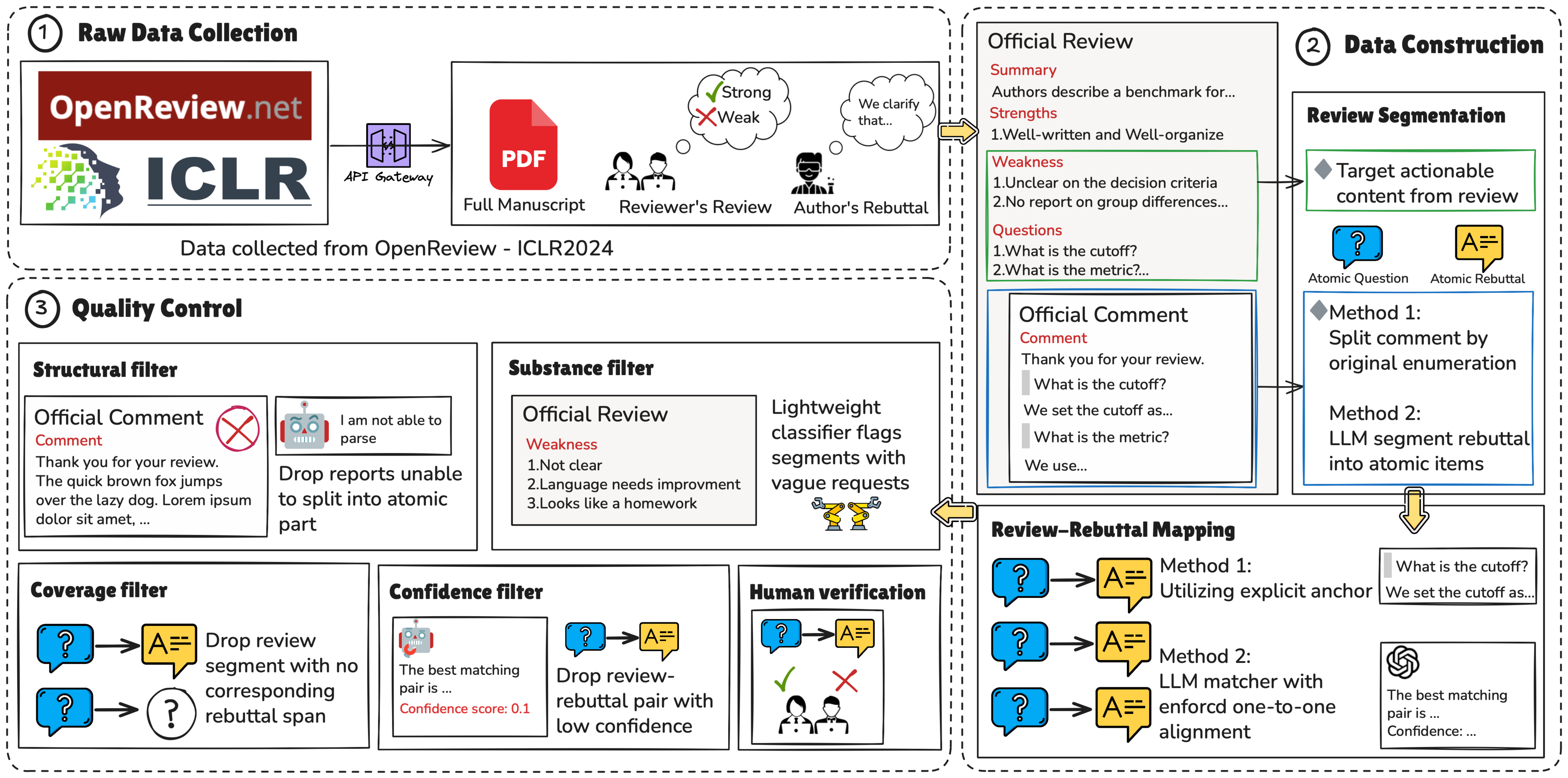}
	\caption{Overview of our construction pipeline for \data.}
	\label{fig:pipeline}
\end{figure*}
\section{Related Work}

\paragraph{Peer Review Datasets.}
Early corpora such as PeerRead and NLPeer collect manuscripts, decisions, and textual reviews to support large-scale analysis of peer review \citep{peerread,nlpeer}.  More recent large-scale resources support fine-tuning reviewer models, including ReviewMT, Review-5K and DeepReview-13K \citep{tan2024peerreviewmultiturnlongcontext,deepreview,cycleresearcher}. Some other resources incorporate rebuttal to study review: PRRCA links reviews with rebuttal counter-arguments for meta-review generation \citep{prrca}; MOPRD aggregates multi-disciplinary open reviews with rebuttal letters and editorial signals \citep{moprd}; Re$^2$ scales multi-turn review and rebuttal discussions across many venues \citep{re2}. To analyze review--rebuttal links, DISAPERE adds sentence-level discourse annotations over 506 review and rebuttal pairs \citep{disapere}; JitsuPeer further aligns review–rebuttal sentences with rebuttal action types for attitude-grounded rebuttal generation \citep{purkayastha2023exploringjiujitsuargumentationwriting}. However, these links are defined at the sentence level and remain small-scale. 
In contrast, we construct large-scale point-to-point mappings that align each review point to its corresponding rebuttal segment and add perspective and impact labels, using them to supervise review feedback generation.

\paragraph{Review Generation.}
Early efforts prompted LLMs to write full reviews, revealing limited but non-trivial usefulness and notable failure modes \citep{yuan2021automatescientificreviewing,reviewergpt,gpt4sli}. Later systems improved specificity through fine-tuning and structured pipelines, and most recently multi-agent approaches that assign roles to reduce generic feedback and increase comment quality \citep{agentreview,deepreview,diag}. Another related thread focuses on \emph{actionable} reviewing, concerning whether feedback induces concrete changes or commitments. Most prior approaches operationalize actionability using reviewer-centric signals. ReAct labels reviews for actionability and intent to support detection and triage \citep{react}; ARIES links review comments to implemented edits, enabling analysis of which feedback leads to revisions \citep{aries}; MARG explicitly frames the task as generating actionable peer review with specialized agents and reports reduced generic comments \cite{marg}. 
In contrast, our method leverages rebuttal-anchored signals to ground actionability in authors' responses and actions, facilitating actionable review feedback generation.


\section{Review and Rebuttal Mapping}
\label{sec:dataset}

To train models that generate actionable review comments using rebuttal as supervision, we need fine-grained mappings that connect each review point to the specific rebuttal that addresses it. Such segment-level alignment lets us attach a perspective label and an impact signal indicating whether the comment led to a concrete revision, a specific plan, or a defense without changes. Existing resources are not sufficient: 
DISAPERE \citep{disapere} provides sentence-level discourse annotations but lacks our segment-level labels and is much smaller (506 pairs); ARIES \citep{aries} is related but targets authors' edits; other datasets either omit rebuttal or do not treat it as a training signal. Therefore we introduce our dataset \data\ (Review-Map-Rebuttal) which maps review and rebuttal at the segment level, and is much bigger (about 150$\times$ more than DISAPERE).



\subsection{Review–Rebuttal Source Data Collection}
We curate a corpus with papers, reviews and rebuttals from ICLR~2024 on OpenReview by querying the official API for each submission's manuscript, full reviews, and the corresponding author responses.
For downstream text processing, we convert the PDF manuscripts to Markdown with MinerU \cite{wang2024mineruopensourcesolutionprecise}. 
For each paper we keep (i) the full manuscript text; (ii) every reviewer's complete review 
and free-form comments; (iii) the author rebuttal thread. When a rebuttal appears in multiple messages, we concatenate them in chronological order to obtain a single response document per paper.

\subsection{Dataset Construction}
\label{sec:mapping}
We construct a point-to-point mapping between reviewer key points and the specific portion of the rebuttal that addresses each point, named \data, shown in~\autoref{fig:pipeline}. 
Dataset statistics are summarized in~\autoref{tab:rrm75k_stats}.
\begin{table}[h]
\centering
\small
\setlength{\tabcolsep}{10pt}
\begin{tabular}{l r}
\toprule
\textbf{Statistic} & \textbf{Value} \\
\midrule
Total mappings & 75{,}542 \\
Total papers & 4{,}825 \\
Avg.\ reviewers per paper & 3.44 \\
Avg.\ mappings per paper & 15.66 \\
Distinct reviews & 16{,}583 \\
Avg.\ mappings per review & 4.56 \\
Avg.\ confidence score & 0.9268 \\
Conference & ICLR 2024 \\
\bottomrule
\end{tabular}
\caption{Summary statistics for the Review-Rebuttal-Mapping-75k dataset.}
\label{tab:rrm75k_stats}
\end{table}

\paragraph{Notation.}
Let paper $p$ have reviewer set $\mathcal{J}_p$. A reviewer $j\!\in\!\mathcal{J}_p$ writes a review $R^{p}_{j}$, which we decompose into a sequence of weakness/question segments $R^{p}_{j,1:K_j} \,{=}\, \{r^{p}_{j,1},\dots,r^{p}_{j,K_j}\}$. The concatenated rebuttal text for $p$ is $B^{p}$, which we further split into candidate response spans $B^{p}_{1:T} \,{=}\, \{b^{p}_{1},\dots,b^{p}_{T}\}$. The goal is to construct a mapping
\[
\mathcal{A}(p) \;=\; \{\, (r^{p}_{j,k},\, b^{p}_{t},\, \hat{c}^{p}_{j,k,t}) \,\}_{(j,k,t)\in\mathcal{I}_p},
\]
where $\hat{c}^{p}_{j,k,t}\in[0,1]$ is a confidence score and we enforce a one-to-one mapping: each $r^{p}_{j,k}$ and each $b^{p}_{t}$ appears in at most one pair.

\paragraph{Review Segmentation.}
We target only actionable content by extracting \emph{Weaknesses} and \emph{Questions} because author rebuttals mainly respond to these two parts. When reviewers already enumerate items (e.g., ``1/2/3'', ``W1/W2/…'' or bullet lists), we take those spans directly. Otherwise, we prompt GPT-5 \cite{gpt5} to split $R^{p}_{j}$ into atomic critique units $\{r^{p}_{j,k}\}$ that each express a single concern. The prompt is shown in \autoref{fig:promptseg}. 


\paragraph{Review–Rebuttal Mapping.}
We perform two-stage alignment. First, a heuristic pass links items using explicit anchors (e.g., "W1", quoted phrases, or reviewer's specific references), yielding high precision pairs. Second, an LLM matcher conducts span-level semantic linking: given a segment $r^{p}_{j,k}$ and the candidate spans $\{b^{p}_{t}\}$, it selects the best-supported $b^{p}_{t}$ and returns a calibrated confidence $\hat{c}^{p}_{j,k,t}$. We enforce one-to-one alignment by greedy matching in descending $\hat{c}$ and discard ties. This realizes a paper-specific map
\[
\mathrm{Map}: \big(R^{p},\, B^{p}\big)\rightarrow \big\{(r^{p}_{j,k},\, b^{p}_{t})\big\}_{(j,k,t)}.
\]
The prompt is shown in \autoref{fig:promptmap}. 


\subsection{Cleaning and Quality Control}
\label{sec:cleaning}
Because our training uses rebuttal-derived impact categories as preference signals for actionability, supervision quality hinges on precise review–rebuttal alignment. We therefore keep only confidently aligned pairs via the filters below and verify quality with targeted human checks.

\smallskip
\noindent\textbf{Structural Filter.} We drop reviews with no visible itemization and for which the segmenter cannot produce stable atomic units. 

\noindent\textbf{Coverage Filter.} During alignment, if a review segment $\;r^{p}_{j,k}\;$ receives no plausible rebuttal span (no $\hat{c}\ge\tau$), we remove it.  

\noindent\textbf{Confidence Filter.} We set a threshold $\tau$ and keep only pairs with $\hat{c}\ge\tau$; we also prune pairs where the matched rebuttal span merely restates the question without addressing it.  

\noindent\textbf{Substance Filter.} We filter out review segments that do not semantically raise any substantive issue or recommendation.

\noindent\textbf{Human Verification.} We sample 60 papers to cover different perspectives. From these papers, we include all review segments that passed our filters and met the confidence threshold, yielding 944 mapped segments. Two trained annotators independently map each review segment to the most specific rebuttal span (or mark ``No Response''), following the same one-to-one constraint and span granularity as our pipeline. A mapping is counted as \emph{correct} if the predicted rebuttal span overlaps the gold span with token-level IoU $\geq 0.5$. We compute span-level Precision/Recall/F1 for our automatic mapping against the adjudicated gold, and report Cohen's~$\kappa$ between the two annotators before adjudication. Results are in \autoref{tab:mapping_validation}. 
Our mapping achieves high span-level accuracy (F1 = 0.91) with substantial IAA ($\kappa=0.80$), indicating that the pipeline provides reliable supervision for downstream training.

\begin{table*}[h]
\centering
\small
\setlength{\tabcolsep}{8pt}
\begin{tabular}{lcccc}
\toprule
\textbf{Setting} & \textbf{Precision} & \textbf{Recall} & \textbf{F1} & \textbf{Cohen's $\kappa$} \\
\midrule
Filtered set & 0.93 & 0.90 & \textbf{0.91} & 0.80 \\
\bottomrule
\end{tabular}
\caption{Gold-standard validation of review$\rightarrow$rebuttal span mappings. $\kappa$ is inter-annotator agreement (IAA).}
\label{tab:mapping_validation}
\end{table*}

\section{Methodology}

\subsection{Task Definition}
\label{sec:taskdef}
Given a paper $p$, a perspective $s$
and the whole paper text as context, the model generates a single focused review segment $y$ that raises weaknesses or questions of paper $p$ from perspective $s$.

\paragraph{Review Perspective Labels.}
Each mapped review segment receives one label from a 7-category taxonomy: \emph{Experiments}, \emph{Writing}, \emph{Presentation}, \emph{Theory}, \emph{Novelty}, \emph{Reproducibility}, and \emph{Evaluation}. We assign labels automatically with a rubric-based prompt to GPT-5 that asks for the single best label and a short rationale, where prompt is shown in Appendix \ref{app:label}.
To verify quality, two trained annotators conducted stratified spot checks over papers and confidence bins. They independently reviewed a held-out sample and then resolved disagreements. The automatic labels matched human judgment on most cases (accuracy about 92\%; Cohen’s $\kappa=0.81$). The remaining confusions were mainly \emph{Writing} vs. \emph{Presentation} and \emph{Experiments} vs. \emph{Evaluation}. Moreover, the detailed definition of perspective labels is shown in~\autoref{tab:labels_detail}.

\paragraph{Rebuttal Impact Category as Actionability Signal.}
For each mapped rebuttal segment, we annotate an impact category that reflects the author's concrete action in response to the review by GPT-5:
(1) \emph{Concrete Revision Performed} (\emph{CRP}), (2) \emph{Specific revision plan} (\emph{SRP}), (3) \emph{Vague Commitment to Revise} (\emph{VCR}), (4) \emph{Defend Without Change} (\emph{DWC}),  and (5) \emph{Deflect/Reframe} (\emph{DRF}). 
The definition is shown in~\autoref{fig:labels_and_composition}, and more detailed definition is shown in~\autoref{tab:labels_detail}.
The automatic labels matched the adjudicated human labels on most cases (accuracy 89\%), with inter-annotator agreement $\kappa=0.79$.
They correspond to varying degrees of modifications made by the authors, reflecting the review's level of actionability.
The categories leverage author reactions in rebuttal segments, reflecting how much revision or defense occurred, to capture the actionability of reviewer comments and their impact on rebuttals. After labeling, the distribution of each impact category and perspective in \data\ is shown in \autoref{fig:labels_and_composition}. We detail the specific label process and the distribution of each category in Appendix \ref{app:label}.

\begin{figure*}[t]
\centering

\begin{minipage}[t]{0.52\linewidth}
\centering
\vspace{0pt}
\small
\setlength{\tabcolsep}{6pt}
\begin{tabular}{ll}
\toprule
\textbf{Label set} & \textbf{Definition} \\
\midrule
\multicolumn{2}{c}{\itshape Perspective of Review Segment} \\
\cmidrule(lr){1-2}
\emph{Experiments} & Setup, baselines, ablations, datasets \\
\emph{Evaluation} & Metrics, analysis, claims vs.\ results \\
\emph{Reproducibility} & Missing code/details, reproducibility info \\
\emph{Novelty} & Originality, relation to prior work \\
\emph{Theory} & Assumptions, derivations, proofs \\
\emph{Writing} & Clarity, grammar, readability \\
\emph{Presentation} & Figures, tables, organization \\
\cmidrule(lr){1-2}
\multicolumn{2}{c}{\itshape Impact Category of Rebuttal Segment} \\
\cmidrule(lr){1-2}
\emph{CRP} & Concrete Revision Performed \\
\emph{SRP} & Specific Revision Plan \\
\emph{VCR} & Vague Commitment to Revise \\
\emph{DWC} & Defend Without Change \\
\emph{DRF} & Deflect or reframe, no change \\
\bottomrule
\end{tabular}
\end{minipage}
\hfill
\begin{minipage}[t]{0.45\linewidth}
\centering
\vspace{0pt}
\begin{tikzpicture}
\begin{axis}[
    ybar stacked,
    bar width=10pt,
    width=\linewidth,
    height=5.8cm,
    ymin=0, ymax=100,
    ylabel={Percentage (\%)},
    symbolic x coords={Evaluation,Experiments,Novelty,Presentation,Reproducibility,Theory,Writing},
    xtick=data,
    xticklabel style={rotate=30, anchor=east, font=\footnotesize, yshift=-3pt},
    legend style={
        at={(0.5,1.03)},
        anchor=south,
        legend columns=5,
        /tikz/every even column/.append style={column sep=4pt}
    },
    enlarge x limits=0.03,
    ymajorgrids,
]

\addplot coordinates {(Evaluation,42.3) (Experiments,47.9) (Novelty,32.9) (Presentation,60.6) (Reproducibility,45.6) (Theory,33.2) (Writing,55.0)};
\addplot coordinates {(Evaluation, 8.0) (Experiments, 9.0) (Novelty,10.2) (Presentation,16.8) (Reproducibility,10.6) (Theory, 8.7) (Writing,13.5)};
\addplot coordinates {(Evaluation, 1.5) (Experiments, 1.6) (Novelty, 2.2) (Presentation, 5.4) (Reproducibility, 2.7) (Theory, 2.2) (Writing, 7.4)};
\addplot coordinates {(Evaluation,46.6) (Experiments,39.1) (Novelty,53.3) (Presentation,16.4) (Reproducibility,39.7) (Theory,53.5) (Writing,23.4)};
\addplot coordinates {(Evaluation, 1.5) (Experiments, 2.4) (Novelty, 1.4) (Presentation, 0.8) (Reproducibility, 1.4) (Theory, 2.5) (Writing, 0.8)};

\legend{CRP, SRP, VCR, DWC, DRF}
\end{axis}
\end{tikzpicture}
\end{minipage}

\caption{\textbf{Left:} brief summary of perspective labels for review segments and impact categories for rebuttal segments. \textbf{Right:} normalized (100\%) impact category composition by perspective.}
\label{fig:labels_and_composition}
\end{figure*}

\subsection{Training Dataset Construction}
\label{sec:sft}

\paragraph{SFT Data.}
We construct \textsc{ReviewSeg-SFT-13K}, a supervised corpus collecting pairs $\mathcal{D}_{\mathrm{SFT}}=\{(x_i,y_i)\}_{i=1}^{N}$.
Each input is $x_i=(p, s)$, where $p$ is the paper content and $s$ is the target perspective (e.g., \emph{Experiments}, \emph{Writing}); the target $y_i^\star$ is the gold review sentence selected from the mapped reviewer segments $R^{p}_{j,1:K_j}$.
Dataset size and coverage are reported in~\autoref{tab:data_sft_dpo}. There are $13{,}300$ pairs spanning $4{,}637$ papers, balanced to include $1{,}900$ instances per perspective.

\paragraph{Preference Data.}
From the review–rebuttal alignments $\mathcal{A}(p)$ in \S\ref{sec:mapping}, we build \textsc{ReviewPref-DPO-22K} as preference triples $(x, y_w, y_\ell)$, where $x=(p,s)$ matches the SFT input and $y_w,y_\ell$ are two review segments drawn from the same paper $p$ and perspective $s$. Winners follow the rebuttal impact order \(\emph{CRP} > \emph{SRP} > \emph{VCR} > \emph{DWC} > \emph{DRF}\), which reflects increasing author uptake and revision degree in rebuttals. We compare only strictly ordered labels and stratify by impact gap to modulate difficulty: large (\emph{CRP} vs.\ {\emph{DWC}, \emph{DRF}}), medium (\emph{SRP} vs.\ {\emph{DWC}, \emph{DRF}}, \emph{CRP} vs.\ \emph{VCR}), and small (\emph{CRP} vs.\ \emph{SRP}, \emph{VCR} vs.\ {\emph{DWC}, \emph{DRF}}, \emph{DWC} vs.\ \emph{DRF}). To keep the signal robust, we balance pair counts across papers and perspectives, never mix different papers or perspectives within a pair, and cap how often any single segment can appear so no segment dominates the pool.
As summarized in~\autoref{tab:data_sft_dpo}, the preference set contains $21{,}822$ pairs from $4{,}825$ papers with $3{,}117$ pairs per perspective in average. 

Moreover, we quantify the difficulty stratification in \autoref{tab:dpo_difficulty}.
This distribution exposes the model to a spectrum of preference margins while keeping pairs within the same paper and perspective.

\begin{table}[h]
\centering
\small
\setlength{\tabcolsep}{5pt}
\begin{tabular}{lrr}
\toprule
\textbf{Statistic} & \textsc{SFT-13k} & \textsc{DPO-22k} \\
\midrule
\#\,pairs & 13{,}300 & 21{,}822 \\
\#\,papers & 4{,}637 & 4{,}825 \\
Per-perspective & $1{,}900$ each & avg. 3{,}117 each \\
Avg. paper length & 22{,}152 & 21{,}798 \\
Avg. output length & 62 & Ch: 65, Re: 63 \\
Conference & ICLR 2024 & ICLR 2024 \\
Extra test pairs (\(\approx\)10\%) & 1330 & 2180 \\
\bottomrule
\end{tabular}
\caption{Statistics for SFT dataset and preference dataset for DPO. The unit of length is tokens. ``Ch'' denotes Chosen and ``Re'' denotes Rejected.}
\label{tab:data_sft_dpo}
\end{table}

\begin{table}[h]
\centering
\small
\setlength{\tabcolsep}{6pt}
\begin{tabular}{lllrr}
\toprule
\textbf{Tier} & \textbf{Winner} & \textbf{Loser} & \textbf{\#Pairs} & \textbf{Percentages} \\
\midrule
\multicolumn{5}{l}{\textit{Large impact gap (Easy)}}\\
& \textsc{CRP} & \textsc{DWC} & 9{,}887 & 45.3 \\
& \textsc{CRP} & \textsc{DRF} & 381 & 1.7 \\
\textit{Subtotal} & & & 10{,}268 & 47.1 \\
\midrule
\multicolumn{5}{l}{\textit{Medium impact gap}}\\
& \textsc{SRP} & \textsc{DWC} & 4{,}969 & 22.8 \\
& \textsc{SRP} & \textsc{DRF} & 248 & 1.1 \\
& \textsc{CRP} & \textsc{VCR} & 1{,}904 & 8.7 \\
\textit{Subtotal} & & & 7{,}121 & 32.6 \\
\midrule
\multicolumn{5}{l}{\textit{Small impact gap (Hard)}}\\
& \textsc{CRP} & \textsc{SRP} & 2{,}484 & 11.4 \\
& \textsc{VCR} & \textsc{DWC} & 1{,}423 & 6.5 \\
& \textsc{VCR} & \textsc{DRF} & 70 & 0.3 \\
& \textsc{DWC} & \textsc{DRF} & 456 & 2.1 \\
\textit{Subtotal} & & & 6{,}433 & 20.3 \\
\midrule
\textbf{Total} & & & 21{,}822 & 100.0 \\
\bottomrule
\end{tabular}
\caption{Difficulty-stratified breakdown of preference pairs in \textsc{ReviewPref-DPO-22k}. Pairs are constructed only within the same paper and perspective and use strictly ordered rebuttal-impact labels.}
\label{tab:dpo_difficulty}
\end{table}

\subsection{Policy Optimization}
\label{sec:dpo}


\paragraph{Direct Preference Optimization.}
Direct Preference Optimization (DPO) learns a policy from pairwise preferences without fitting a separate reward model \citep{dpo}. Let $\pi_{\theta}$ be the trainable policy and $\pi_{\mathrm{ref}}$ a fixed reference policy (the SFT model). Under a Bradley–Terry preference model, DPO maximizes the probability that the policy prefers $y_{w}$ over $y_{\ell}$ by matching the optimal policy’s log-density ratio relative to $\pi_{\mathrm{ref}}$. The loss for a batch of preference triples is as follows:
Let $\Delta_{\theta,\mathrm{ref}}(x,y) = 
\log \pi_{\theta}(y\!\mid\!x)-\log \pi_{\mathrm{ref}}(y\!\mid\!x)$.
We sample preference triples $(x,y_w,y_\ell)$ from the dataset 
$\mathcal{D}_{\mathrm{pref}}$. The DPO loss is
\begin{equation}
\resizebox{0.6\linewidth}{!}{
$
\mathcal{L}_{\mathrm{DPO}}(\theta)
= -\;\mathbb{E}_{(x,y_w,y_\ell)} \Big[
\log \sigma\!\big(
\beta \,[\,\Delta_{\theta,\mathrm{ref}}(x,y_w) - \Delta_{\theta,\mathrm{ref}}(x,y_\ell)\,]
\big)\Big]
$}
\end{equation}

\noindent where $\beta\!>\!0$ controls the sharpness of the preference. Intuitively, the policy is encouraged to increase likelihood on comments that led to higher-impact author actions (\textsc{CRP}, \textsc{SRP}) while decreasing it on comments that were defended or deflected (\textsc{DWC}, \textsc{DRF}).

\paragraph{Stabilization.}
We keep $\pi_{\mathrm{ref}}$ frozen at the SFT checkpoint and optionally mix in a small fraction of $\mathcal{L}_{\mathrm{SFT}}$ on positive samples to prevent drift on perspective control when the context is long. The full objective is
\[
\mathcal{L}(\theta)\,=\,\mathcal{L}_{\mathrm{DPO}}(\theta)\;+\;\lambda\,\mathbb{E}_{(x,y_w)}\!\left[-\log \pi_{\theta}(y_w\!\mid\!x)\right],
\]
with a small $\lambda$. In practice, we set $\lambda$ as 0.1.




\subsection{Training Details}
\label{sec:train_details}

All experiments run on NVIDIA H200 141GB GPUs with bf16 compute. 
We use the base model Llama-3.1-8B-Instruct \cite{llama31}.
We train \textsc{RbtAct}-SFT model on dataset \textsc{ReviewSeg-SFT-13k}.
We run $3$ epochs ($4{,}989$ steps) with learning rate $1.0{\times}10^{-4}$ and a cosine scheduler.
Total training time on H200 is approximately $120$ hours.
To obtain \textsc{RbtAct}, we then utilize dataset \textsc{ReviewPref-DPO-22k} for further DPO training.
The DPO policy $\pi_\theta$ is initialized from the SFT checkpoint and trained for $2$ epochs ($2{,}728$ steps) using the Bradley–Terry DPO loss, keeping the reference $\pi_{\text{ref}}$ frozen.
We use learning rate $1.0{\times}10^{-5}$ and a cosine scheduler.
Total DPO training time on H200 is approximately $203$ hours.
The additional training details including training configs and other optimizations are shown in Appendix~\ref{app:train_details}.

\section{Experiments}
\label{sec:exp}


\subsection{Baselines}
We compare \textsc{RbtAct} against three baseline types under identical inputs, prompts and decoding.

\paragraph{Fine-tuned (SFT-only).}
As a trained baseline, we fine-tune Llama-3.1-8B-Instruct on ReviewSeg-SFT-13k, getting \textsc{RbtAct}-SFT, using the same prompt template and output format.

\paragraph{Prompted LLMs.}
We query API models and run competitive open-source models in zero-shot mode to produce a single review segment for the requested perspective:
GPT-5-chat \cite{gpt5}, DeepSeek-V3.2 \cite{deepseekv32}, Llama-3.1-70B \cite{llama31} and Qwen-3-32B \cite{qwen32}. The prompt is shown in Appendix \ref{app:llmasjudgeprompt}.

\paragraph{Other Methods.}
We additionally compare against three task-adapted methods that are widely used around review feedback generation: (i) MARG~\cite{marg}, a multi-agent prompting framework for scientific review generation; (ii) LimGen~\cite{limgen}, a limitations-focused generation setup that targets suggestive weaknesses; and (iii) DeepReviewer-14B~\cite{deepreview}, a review generation model to produce comprehensive reviews. For each method, we adapt its protocol to our setting (paper + requested perspective $\rightarrow$ a single review feedback) and normalize the final output to our unified segment format. The details are shown in \ref{app:baseline_adapt}.



\subsection{Evaluation Protocol}
\paragraph{Evaluation Dataset.}
We construct a test set from a subset of ICLR 2025 papers using the same pipeline as in \S\ref{sec:mapping} to reduce data contamination, ensuring that none are resubmissions of ICLR 2024 papers. We sample 700 papers, stratified by perspective with 100 papers per perspective across the seven perspectives. Each paper is paired with one perspective labeled by annotators and one human review segment, which serves as the gold reference. 

\paragraph{Human Evaluation Setup.}

We evaluate a subset of the evaluation dataset: 50 papers sampled across perspectives. For each paper, annotators view the title, relevant content, and the target perspective. Three PhD-level or senior graduate annotators (each with $\geq$2 completed reviews at major ML venues) rate nine anonymized model outputs per paper on a 1–5 scale for \emph{Actionability}, \emph{Specificity}, \emph{Groundedness}, \emph{Relevance}, and \emph{Helpfulness}, following a written rubric with positive and negative indicators and anchor examples. Outputs are shown as three order-randomized pairs to mitigate position bias. Scores are averaged over annotators. Full instructions are reported in Appendix~\ref{app:humaneval}.

\paragraph{LLM-as-a-Judge Evaluation Setup.}

We evaluate on 105 papers from the evaluation dataset. Each model produces one segment per paper, yielding 105 segments per model and 945 total across nine models. The judge model (GPT-5-chat) sees the paper context, the target perspective, and one or two candidate segments \cite{zheng2023judgingllmasajudgemtbenchchatbot}. For pointwise scoring, it assigns 1–5 on the same five dimensions as the human study. For pairwise comparison, it scores both candidates and then chooses which is more \emph{Actionable} with a brief rationale. The exact prompts are in \autoref{fig:promptllmpoint} and \autoref{fig:promptllmpair}.

\begin{table*}[h]
\centering
\small
\setlength{\tabcolsep}{3.2pt}
\renewcommand{\arraystretch}{1.08}
\begin{tabular}{lccccc@{\hspace{10pt}}ccccc}
\toprule
& \multicolumn{5}{c}{\textbf{Human}} & \multicolumn{5}{c}{\textbf{LLM-as-a-Judge}} \\
\cmidrule(r){2-6} \cmidrule(l){7-11}
\textbf{System} & \textbf{Action.} & \textbf{Spec.} & \textbf{Ground.} & \textbf{Rel.} & \textbf{Help.}
               & \textbf{Action.} & \textbf{Spec.} & \textbf{Ground.} & \textbf{Rel.} & \textbf{Help.} \\
\midrule

\rowcolor{grpours}
\multicolumn{11}{c}{\textbf{Ours}} \\
\textsc{RbtAct} (ours) 
& \textbf{3.46} & \textbf{4.08} & 4.30 & 4.76 & 4.26
& \textbf{3.38} & \textbf{3.70} & 4.05 & 4.82 & 3.74 \\
\textsc{RbtAct}-SFT    
& 3.28 & 4.01 & 4.16 & 4.70 & 4.24
& 3.18 & 3.59 & 3.94 & 4.72 & 3.66 \\
\addlinespace[2pt]

\rowcolor{grpllm}
\multicolumn{11}{c}{\textbf{LLMs}} \\
GPT-5-chat      
& 3.38 & 4.04 & \textbf{4.35} & \textbf{4.98} & \textbf{4.47}
& 3.28 & 3.66 & \textbf{4.12} & \textbf{4.95} & \textbf{3.78} \\
DeepSeek-V3.2   
& 3.15 & 3.98 & 4.22 & 4.88 & 4.28
& 3.13 & 3.56 & 4.00 & 4.86 & 3.70 \\
Llama-3.1-70B   
& 3.22 & 3.95 & 4.18 & 4.65 & 4.15
& 3.11 & 3.54 & 3.96 & 4.74 & 3.53 \\
Qwen-3-32B      
& 3.06 & 3.78 & 4.12 & 4.58 & 4.12
& 3.03 & 3.36 & 3.90 & 4.68 & 3.32 \\
\addlinespace[2pt]

\rowcolor{grpother}
\multicolumn{11}{c}{\textbf{Other Methods}} \\
MARG            
& 3.20 & 3.87 & 4.15 & 4.72 & 4.18
& 3.19 & 3.47 & 3.91 & 4.69 & 3.51 \\
DeepReviewer-14B  
& 3.27 & 3.96 & 4.28 & 4.75 & 4.21
& 3.23 & 3.48 & 4.03 & 4.74 & 3.58 \\
LimGen          
& 3.14 & 3.92 & 4.08 & 4.64 & 4.05
& 3.08 & 3.38 & 3.88 & 4.54 & 3.39 \\
\bottomrule
\end{tabular}
\caption{Pointwise ratings on five quality dimensions. Left: Human. Right: LLM-as-a-judge. 
Higher is better.}
\label{tab:pointwise-combined}
\end{table*}

\subsection{Experiment Results}
\label{sec:expresults}
\paragraph{Human Evaluation Results.}
As shown in \autoref{tab:pointwise-combined}, \textsc{RbtAct} attains the highest \emph{Actionability} while remaining competitive on other dimensions.

\paragraph{LLM-as-a-judge Pointwise Results.}
The judge reproduces the human trend: \textsc{RbtAct} scores highest on \emph{Actionability} and \emph{Specificity}, while remaining close to strong LLM baselines on other dimensions (\autoref{tab:pointwise-combined}). The prompt is shown in \autoref{fig:promptllmpoint}.

\paragraph{LLM-as-a-judge Pairwise Results.}
We compare \emph{Actionability} via pairwise judgments on the same paper and perspective. 
\autoref{tab:llm-pairwise} reports the percentage of wins for each model. 
To visualize patterns across perspectives, \autoref{fig:hm_overall} shows heatmaps where each cell is the win rate of the \emph{row} model over the \emph{column} model. Overall, \textsc{RbtAct} attains the highest average pairwise win rate and leads in most perspectives, followed by GPT-5-chat. The pairwise prompt is in \autoref{fig:promptllmpair}.

\begin{table*}[h]
\centering
\small
\setlength{\tabcolsep}{4pt}
\renewcommand{\arraystretch}{0.9}
\resizebox{\linewidth}{!}{%
\begin{tabular}{lccccccccc}
\toprule
& \multicolumn{8}{c}{\textbf{Actionability (Pairwise Win Rate \%)}} \\
\cmidrule(lr){2-10}
Winners & \textsc{RbtAct} & GPT-5 & DeepSeek & Llama-3.1 & DeepReviewer & MARG & Qwen-3 & \textsc{RbtAct}-SFT & LimGen \\
\midrule
\textsc{RbtAct}      & --   & 57.1 & 63.8 & 61.9 & 65.7 & 68.6 & 66.7 & 65.7 & 76.2 \\
GPT-5-chat           & 42.9 & --   & 55.2 & 57.1 & 60.0 & 62.9 & 60.0 & 54.3 & 71.4 \\
DeepSeek-V3.2        & 36.2 & 44.8 & --   & 54.3 & 57.1 & 59.0 & 58.1 & 53.3 & 68.6 \\
Llama-3.1-70B        & 38.1 & 42.9 & 45.7 & --   & 52.4 & 55.2 & 56.2 & 51.4 & 66.7 \\
DeepReviewer-14B       & 34.3 & 40.0 & 42.9 & 47.6 & --   & 51.4 & 53.3 & 50.5 & 64.8 \\
MARG                 & 31.4 & 37.1 & 41.0 & 44.8 & 48.6 & --   & 50.5 & 49.5 & 61.9 \\
Qwen-3-32B           & 33.3 & 40.0 & 41.9 & 43.8 & 46.7 & 49.5 & --   & 52.4 & 58.1 \\
\textsc{RbtAct}-SFT  & 34.3 & 45.7 & 46.7 & 48.6 & 47.6 & 50.5 & 47.6 & --   & 56.2 \\
LimGen               & 23.8 & 28.6 & 31.4 & 33.3 & 35.2 & 38.1 & 41.9 & 43.8 & --   \\
\bottomrule
\end{tabular}
}
\caption{LLM-as-a-judge pairwise win rates (\%) on Actionability}
\label{tab:llm-pairwise}
\end{table*}

\paragraph{Human–LLM Agreement.}
\label{sec:human-llm-agree}
We quantify alignment between human judgments and the LLM judge along three axes.
(i) \emph{Model-level rank correlation}: ranking models by pointwise means yields high agreement for \emph{Actionability} (Spearman’s $\rho{=}0.94$, Kendall’s $\tau_b{=}0.87$), with only a minor swap between mid-ranked baselines.
(ii) \emph{Item-level correlation}: across matched paper–model cells (20 papers $\times$ 9 models), human vs.\ LLM pointwise scores show moderate positive correlation on \emph{Actionability} (Spearman’s $\rho{=}0.52$).

\subsection{Automatic Evaluation}
\label{sec:auto-eval}
We evaluate all 700 instances from evaluation dataset with four metrics: BLEU@4 \cite{papineni-etal-2002-bleu}, ROUGE-L\textsubscript{sum} (F1) \cite{lin-2004-rouge}, METEOR \cite{banerjee-lavie-2005-meteor}, and chrF \cite{popovic-2015-chrf}. Details are shown in Appendix \ref{app:autoeval}.
Our models are competitive with or stronger than the baselines on most overlap-based metrics. In particular, \textsc{RbtAct} achieves the best ROUGE-L\textsubscript{sum} and METEOR, while \textsc{RbtAct}-SFT performs similarly and attains the highest BLEU@4.




\subsection{Case Study}
Here we show some case studies to demonstrate why our model \textsc{RbtAct} is more actionable than other baselines in most cases from different perspectives. They are shown in \autoref{app:casestudy}.


\subsection{Summary}
Both human and LLM-as-a-judge evaluations show that our \textsc{RbtAct} improves the practical value of review segments. Automatic evaluation also shows the strong capability of \textsc{RbtAct}. Gains concentrate on actionability and specificity with parity on groundedness, relevance, and Helpfulness. Importantly, our 8B-scale \textsc{RbtAct} remains competitive with much larger 32–70B and proprietary models on actionability (and often specificity), suggesting the practical value of rebuttal-supervised training. Additional analyses on issue severity and paper strength are provided in Appendix~\ref{app:discuss}, showing that our gains are not driven only by minor issues and are also consistent across papers of different strengths.


\section{Conclusion}
We study actionable review feedback generation by placing rebuttal at the center of learning. Our framework, \textsc{RbtAct}, uses rebuttals as implicit supervision and frames the task as segment-level generation from a given perspective with explicit mappings from each review segment to its addressing rebuttal span. We release \data\ with perspective labels and impact categories that utilize actionability. An effective pipeline of supervised fine-tuning followed by preference optimization on impact ordered pairs yields consistent gains in actionability and specificity while maintaining grounding and relevance against strong baselines under both human evaluation and LLM-as-a-judge protocols. These results show rebuttal signals are a practical form of human feedback for producing targeted, implementable guidance. We release data and code to spur research on rebuttal-driven learning and better evaluation of actionability.



\section*{Limitations}
Our approach relies on rebuttal as an implicit supervision signal for actionability, which is informative but imperfect. Rebuttals reflect short horizon author uptake during review, not long-term implementation, and can include strategic promises or deferrals. The dataset focuses on venues with public rebuttals, mainly computer science communities that use OpenReview, so generalization to journals, non-English venues, and other fields remains uncertain. Our model can generate precise yet infeasible suggestions and our current setup does not include rigorous verification against the manuscript, code, or data artifacts.

\bibliography{custom}

@inproceedings{peerread,
    title = "A Dataset of Peer Reviews ({P}eer{R}ead): Collection, Insights and {NLP} Applications",
    author = "Kang, Dongyeop  and
      Ammar, Waleed  and
      Dalvi, Bhavana  and
      van Zuylen, Madeleine  and
      Kohlmeier, Sebastian  and
      Hovy, Eduard  and
      Schwartz, Roy",
    editor = "Walker, Marilyn  and
      Ji, Heng  and
      Stent, Amanda",
    booktitle = "Proceedings of the 2018 Conference of the North {A}merican Chapter of the Association for Computational Linguistics: Human Language Technologies, Volume 1 (Long Papers)",
    month = jun,
    year = "2018",
    address = "New Orleans, Louisiana",
    publisher = "Association for Computational Linguistics",
    url = "https://aclanthology.org/N18-1149/",
    doi = "10.18653/v1/N18-1149",
    pages = "1647--1661"
}

@misc{wang2024mineruopensourcesolutionprecise,
      title={MinerU: An Open-Source Solution for Precise Document Content Extraction}, 
      author={Bin Wang and Chao Xu and Xiaomeng Zhao and Linke Ouyang and Fan Wu and Zhiyuan Zhao and Rui Xu and Kaiwen Liu and Yuan Qu and Fukai Shang and Bo Zhang and Liqun Wei and Zhihao Sui and Wei Li and Botian Shi and Yu Qiao and Dahua Lin and Conghui He},
      year={2024},
      eprint={2409.18839},
      archivePrefix={arXiv},
      primaryClass={cs.CV},
      url={https://arxiv.org/abs/2409.18839}, 
}

@misc{dpo,
      title={Direct Preference Optimization: Your Language Model is Secretly a Reward Model}, 
      author={Rafael Rafailov and Archit Sharma and Eric Mitchell and Stefano Ermon and Christopher D. Manning and Chelsea Finn},
      year={2024},
      eprint={2305.18290},
      archivePrefix={arXiv},
      primaryClass={cs.LG},
      url={https://arxiv.org/abs/2305.18290}, 
}

@misc{reviewergpt,
      title={ReviewerGPT? An Exploratory Study on Using Large Language Models for Paper Reviewing}, 
      author={Ryan Liu and Nihar B. Shah},
      year={2023},
      eprint={2306.00622},
      archivePrefix={arXiv},
      primaryClass={cs.CL},
      url={https://arxiv.org/abs/2306.00622}, 
}

@misc{gpt4sli,
      title={GPT4 is Slightly Helpful for Peer-Review Assistance: A Pilot Study}, 
      author={Zachary Robertson},
      year={2023},
      eprint={2307.05492},
      archivePrefix={arXiv},
      primaryClass={cs.HC},
      url={https://arxiv.org/abs/2307.05492}, 
}

@misc{blindspot,
      title={Mind the Blind Spots: A Focus-Level Evaluation Framework for LLM Reviews}, 
      author={Hyungyu Shin and Jingyu Tang and Yoonjoo Lee and Nayoung Kim and Hyunseung Lim and Ji Yong Cho and Hwajung Hong and Moontae Lee and Juho Kim},
      year={2025},
      eprint={2502.17086},
      archivePrefix={arXiv},
      primaryClass={cs.CL},
      url={https://arxiv.org/abs/2502.17086}, 
}

@article{hosseini2023chatgptpeerreview,
  title        = {Fighting reviewer fatigue or amplifying bias? Considerations and recommendations for use of ChatGPT and other large language models in scholarly peer review},
  author       = {Hosseini, M. and Horbach, S.~P.~J.~M.},
  journal      = {Research Integrity and Peer Review},
  volume       = {8},
  number       = {4},
  pages        = {1--12},
  year         = {2023},
  publisher    = {Springer},
  doi          = {10.1186/s41073-023-00133-5},
  url          = {https://doi.org/10.1186/s41073-023-00133-5}
}

@inproceedings{agentreview,
    title = "{A}gent{R}eview: Exploring Peer Review Dynamics with {LLM} Agents",
    author = "Jin, Yiqiao  and
      Zhao, Qinlin  and
      Wang, Yiyang  and
      Chen, Hao  and
      Zhu, Kaijie  and
      Xiao, Yijia  and
      Wang, Jindong",
    editor = "Al-Onaizan, Yaser  and
      Bansal, Mohit  and
      Chen, Yun-Nung",
    booktitle = "Proceedings of the 2024 Conference on Empirical Methods in Natural Language Processing",
    month = nov,
    year = "2024",
    address = "Miami, Florida, USA",
    publisher = "Association for Computational Linguistics",
    url = "https://aclanthology.org/2024.emnlp-main.70/",
    doi = "10.18653/v1/2024.emnlp-main.70",
    pages = "1208--1226"
}

@misc{tan2024peerreviewmultiturnlongcontext,
      title={Peer Review as A Multi-Turn and Long-Context Dialogue with Role-Based Interactions}, 
      author={Cheng Tan and Dongxin Lyu and Siyuan Li and Zhangyang Gao and Jingxuan Wei and Siqi Ma and Zicheng Liu and Stan Z. Li},
      year={2024},
      eprint={2406.05688},
      archivePrefix={arXiv},
      primaryClass={cs.CL},
      url={https://arxiv.org/abs/2406.05688}, 
}

@misc{reviewer2,
      title={Reviewer2: Optimizing Review Generation Through Prompt Generation}, 
      author={Zhaolin Gao and Kianté Brantley and Thorsten Joachims},
      year={2024},
      eprint={2402.10886},
      archivePrefix={arXiv},
      primaryClass={cs.CL},
      url={https://arxiv.org/abs/2402.10886}, 
}

@inproceedings{disapere,
    title = "{DISAPERE}: A Dataset for Discourse Structure in Peer Review Discussions",
    author = "Kennard, Neha Nayak  and
      O{'}Gorman, Tim  and
      Das, Rajarshi  and
      Sharma, Akshay  and
      Bagchi, Chhandak  and
      Clinton, Matthew  and
      Yelugam, Pranay Kumar  and
      Zamani, Hamed  and
      McCallum, Andrew",
    editor = "Carpuat, Marine  and
      de Marneffe, Marie-Catherine  and
      Meza Ruiz, Ivan Vladimir",
    booktitle = "Proceedings of the 2022 Conference of the North American Chapter of the Association for Computational Linguistics: Human Language Technologies",
    month = jul,
    year = "2022",
    address = "Seattle, United States",
    publisher = "Association for Computational Linguistics",
    url = "https://aclanthology.org/2022.naacl-main.89/",
    doi = "10.18653/v1/2022.naacl-main.89",
    pages = "1234--1249"
}

@misc{re2,
      title={Re$^2$: A Consistency-ensured Dataset for Full-stage Peer Review and Multi-turn Rebuttal Discussions}, 
      author={Daoze Zhang and Zhijian Bao and Sihang Du and Zhiyi Zhao and Kuangling Zhang and Dezheng Bao and Yang Yang},
      year={2025},
      eprint={2505.07920},
      archivePrefix={arXiv},
      primaryClass={cs.CL},
      url={https://arxiv.org/abs/2505.07920}, 
}

@misc{aries,
      title={ARIES: A Corpus of Scientific Paper Edits Made in Response to Peer Reviews}, 
      author={Mike D'Arcy and Alexis Ross and Erin Bransom and Bailey Kuehl and Jonathan Bragg and Tom Hope and Doug Downey},
      year={2024},
      eprint={2306.12587},
      archivePrefix={arXiv},
      primaryClass={cs.CL},
      url={https://arxiv.org/abs/2306.12587}, 
}

@inproceedings{swift,
    title = "Automated Focused Feedback Generation for Scientific Writing Assistance",
    author = "Chamoun, Eric  and
      Schlichtkrull, Michael  and
      Vlachos, Andreas",
    editor = "Ku, Lun-Wei  and
      Martins, Andre  and
      Srikumar, Vivek",
    booktitle = "Findings of the Association for Computational Linguistics: ACL 2024",
    month = aug,
    year = "2024",
    address = "Bangkok, Thailand",
    publisher = "Association for Computational Linguistics",
    url = "https://aclanthology.org/2024.findings-acl.580/",
    doi = "10.18653/v1/2024.findings-acl.580",
    pages = "9742--9763"
}

@misc{deepreview,
      title={DeepReview: Improving LLM-based Paper Review with Human-like Deep Thinking Process}, 
      author={Minjun Zhu and Yixuan Weng and Linyi Yang and Yue Zhang},
      year={2025},
      eprint={2503.08569},
      archivePrefix={arXiv},
      primaryClass={cs.CL},
      url={https://arxiv.org/abs/2503.08569}, 
}

@misc{yuan2021automatescientificreviewing,
      title={Can We Automate Scientific Reviewing?}, 
      author={Weizhe Yuan and Pengfei Liu and Graham Neubig},
      year={2021},
      eprint={2102.00176},
      archivePrefix={arXiv},
      primaryClass={cs.CL},
      url={https://arxiv.org/abs/2102.00176}, 
}

@inbook{react,
   title={ReAct: A Review Comment Dataset for Actionability (and more)},
   ISBN={9783030915605},
   ISSN={1611-3349},
   url={http://dx.doi.org/10.1007/978-3-030-91560-5_24},
   DOI={10.1007/978-3-030-91560-5_24},
   booktitle={Web Information Systems Engineering – WISE 2021},
   publisher={Springer International Publishing},
   author={Choudhary, Gautam and Modani, Natwar and Maurya, Nitish},
   year={2021},
   pages={336–343} }

@misc{marg,
      title={MARG: Multi-Agent Review Generation for Scientific Papers}, 
      author={Mike D'Arcy and Tom Hope and Larry Birnbaum and Doug Downey},
      year={2024},
      eprint={2401.04259},
      archivePrefix={arXiv},
      primaryClass={cs.CL},
      url={https://arxiv.org/abs/2401.04259}, 
}

@inproceedings{prrca,
author = {Wu, Po-Cheng and Yen, An-Zi and Huang, Hen-Hsen and Chen, Hsin-Hsi},
title = {Incorporating Peer Reviews and Rebuttal Counter-Arguments for Meta-Review Generation},
year = {2022},
isbn = {9781450392365},
publisher = {Association for Computing Machinery},
address = {New York, NY, USA},
url = {https://doi.org/10.1145/3511808.3557360},
doi = {10.1145/3511808.3557360},
booktitle = {Proceedings of the 31st ACM International Conference on Information \& Knowledge Management},
pages = {2189–2198},
numpages = {10},
location = {Atlanta, GA, USA},
series = {CIKM '22}
}

@article{moprd,
   title={MOPRD: A multidisciplinary open peer review dataset},
   volume={35},
   ISSN={1433-3058},
   url={http://dx.doi.org/10.1007/s00521-023-08891-5},
   DOI={10.1007/s00521-023-08891-5},
   number={34},
   journal={Neural Computing and Applications},
   publisher={Springer Science and Business Media LLC},
   author={Lin, Jialiang and Song, Jiaxin and Zhou, Zhangping and Chen, Yidong and Shi, Xiaodong},
   year={2023},
   month=sep, pages={24191–24206} }

@inproceedings{nlpeer,
    title = "{NLP}eer: A Unified Resource for the Computational Study of Peer Review",
    author = "Dycke, Nils  and
      Kuznetsov, Ilia  and
      Gurevych, Iryna",
    editor = "Rogers, Anna  and
      Boyd-Graber, Jordan  and
      Okazaki, Naoaki",
    booktitle = "Proceedings of the 61st Annual Meeting of the Association for Computational Linguistics (Volume 1: Long Papers)",
    month = jul,
    year = "2023",
    address = "Toronto, Canada",
    publisher = "Association for Computational Linguistics",
    url = "https://aclanthology.org/2023.acl-long.277/",
    doi = "10.18653/v1/2023.acl-long.277",
    pages = "5049--5073"
}

@misc{zheng2023judgingllmasajudgemtbenchchatbot,
      title={Judging LLM-as-a-Judge with MT-Bench and Chatbot Arena}, 
      author={Lianmin Zheng and Wei-Lin Chiang and Ying Sheng and Siyuan Zhuang and Zhanghao Wu and Yonghao Zhuang and Zi Lin and Zhuohan Li and Dacheng Li and Eric P. Xing and Hao Zhang and Joseph E. Gonzalez and Ion Stoica},
      year={2023},
      eprint={2306.05685},
      archivePrefix={arXiv},
      primaryClass={cs.CL},
      url={https://arxiv.org/abs/2306.05685}, 
}

@misc{gpt5,
  title        = {GPT-5},
  author       = {{OpenAI}},
  year         = {2025},
  howpublished = {\url{https://openai.com/gpt-5/}},
  note         = {Accessed: 2025-10-07}
}

@misc{deepseekv32,
      title={DeepSeek-V3.2: Pushing the Frontier of Open Large Language Models}, 
      author={DeepSeek-AI},
      year={2025},
}

@misc{llama31,
  title        = {Introducing Llama 3.1: Our most capable models to date},
  author       = {{Meta AI}},
  year         = {2024},
  howpublished = {\url{https://ai.meta.com/blog/meta-llama-3-1/}},
  note         = {Accessed: 2025-10-07}
}

@article{qwen32,
  title   = {Qwen3 Technical Report},
  author  = {Qwen Team},
  journal = {arXiv preprint arXiv:2505.09388},
  year    = {2025},
  url     = {https://arxiv.org/abs/2505.09388}
}

@article{ghosal2022peer,
  title={Peer review analyze: A novel benchmark resource for computational analysis of peer reviews},
  author={Ghosal, Tirthankar and Kumar, Shubham and Bharti, Prasenjit Kumar and Ekbal, Asif},
  journal={PLOS ONE},
  volume={17},
  number={1},
  pages={e0259238},
  year={2022},
  publisher={Public Library of Science},
  doi={10.1371/journal.pone.0259238},
  pmid={35085252},
  pmcid={PMC8794172}
}

@misc{purkayastha2023exploringjiujitsuargumentationwriting,
      title={Exploring Jiu-Jitsu Argumentation for Writing Peer Review Rebuttals}, 
      author={Sukannya Purkayastha and Anne Lauscher and Iryna Gurevych},
      year={2023},
      eprint={2311.03998},
      archivePrefix={arXiv},
      primaryClass={cs.CL},
      url={https://arxiv.org/abs/2311.03998}, 
}

@misc{cycleresearcher,
      title={CycleResearcher: Improving Automated Research via Automated Review}, 
      author={Yixuan Weng and Minjun Zhu and Guangsheng Bao and Hongbo Zhang and Jindong Wang and Yue Zhang and Linyi Yang},
      year={2025},
      eprint={2411.00816},
      archivePrefix={arXiv},
      primaryClass={cs.CL},
      url={https://arxiv.org/abs/2411.00816}, 
}

@misc{sadallah2025goodbadconstructiveautomatically,
      title={The Good, the Bad and the Constructive: Automatically Measuring Peer Review's Utility for Authors}, 
      author={Abdelrahman Sadallah and Tim Baumgärtner and Iryna Gurevych and Ted Briscoe},
      year={2025},
      eprint={2509.04484},
      archivePrefix={arXiv},
      primaryClass={cs.CL},
      url={https://arxiv.org/abs/2509.04484}, 
}

@misc{limgen,
      title={LimGen: Probing the LLMs for Generating Suggestive Limitations of Research Papers}, 
      author={Abdur Rahman Bin Md Faizullah and Ashok Urlana and Rahul Mishra},
      year={2024},
      eprint={2403.15529},
      archivePrefix={arXiv},
      primaryClass={cs.CL},
      url={https://arxiv.org/abs/2403.15529}, 
}

@inproceedings{xu-etal-2025-llms-identify,
    title = "Can {LLM}s Identify Critical Limitations within Scientific Research? A Systematic Evaluation on {AI} Research Papers",
    author = "Xu, Zhijian  and
      Zhao, Yilun  and
      Patwardhan, Manasi  and
      Vig, Lovekesh  and
      Cohan, Arman",
    editor = "Che, Wanxiang  and
      Nabende, Joyce  and
      Shutova, Ekaterina  and
      Pilehvar, Mohammad Taher",
    booktitle = "Proceedings of the 63rd Annual Meeting of the Association for Computational Linguistics (Volume 1: Long Papers)",
    month = jul,
    year = "2025",
    address = "Vienna, Austria",
    publisher = "Association for Computational Linguistics",
    url = "https://aclanthology.org/2025.acl-long.1009/",
    doi = "10.18653/v1/2025.acl-long.1009",
    pages = "20652--20706",
    ISBN = "979-8-89176-251-0"
}

@inproceedings{
zhao2025sciarena,
title={SciArena: An Open Evaluation Platform for Non-Verifiable Scientific Literature-Grounded Tasks},
author={Yilun Zhao and Kaiyan Zhang and Tiansheng Hu and Sihong Wu and Ronan Le Bras and Yixin Liu and Xiangru Tang and Joseph Chee Chang and Jesse Dodge and Jonathan Bragg and Chen Zhao and Hannaneh Hajishirzi and Doug Downey and Arman Cohan},
booktitle={The Thirty-ninth Annual Conference on Neural Information Processing Systems Datasets and Benchmarks Track},
year={2025},
url={https://openreview.net/forum?id=am6RR85mnc}
}

@inproceedings{zhu-etal-2025-deepreview,
    title = "{D}eep{R}eview: Improving {LLM}-based Paper Review with Human-like Deep Thinking Process",
    author = "Zhu, Minjun  and
      Weng, Yixuan  and
      Yang, Linyi  and
      Zhang, Yue",
    editor = "Che, Wanxiang  and
      Nabende, Joyce  and
      Shutova, Ekaterina  and
      Pilehvar, Mohammad Taher",
    booktitle = "Proceedings of the 63rd Annual Meeting of the Association for Computational Linguistics (Volume 1: Long Papers)",
    month = jul,
    year = "2025",
    address = "Vienna, Austria",
    publisher = "Association for Computational Linguistics",
    url = "https://aclanthology.org/2025.acl-long.1420/",
    doi = "10.18653/v1/2025.acl-long.1420",
    pages = "29330--29355",
    ISBN = "979-8-89176-251-0"
}

@misc{diag,
      title={DIAGPaper: Diagnosing Valid and Specific Weaknesses in Scientific Papers via Multi-Agent Reasoning}, 
      author={Zhuoyang Zou and Abolfazl Ansari and Delvin Ce Zhang and Dongwon Lee and Wenpeng Yin},
      year={2026},
      eprint={2601.07611},
      archivePrefix={arXiv},
      primaryClass={cs.AI},
      url={https://arxiv.org/abs/2601.07611}, 
}

@inproceedings{papineni-etal-2002-bleu,
    title = "{B}leu: a Method for Automatic Evaluation of Machine Translation",
    author = "Papineni, Kishore  and
      Roukos, Salim  and
      Ward, Todd  and
      Zhu, Wei-Jing",
    editor = "Isabelle, Pierre  and
      Charniak, Eugene  and
      Lin, Dekang",
    booktitle = "Proceedings of the 40th Annual Meeting of the Association for Computational Linguistics",
    month = jul,
    year = "2002",
    address = "Philadelphia, Pennsylvania, USA",
    publisher = "Association for Computational Linguistics",
    url = "https://aclanthology.org/P02-1040/",
    doi = "10.3115/1073083.1073135",
    pages = "311--318"
}

@inproceedings{lin-2004-rouge,
    title = "{ROUGE}: A Package for Automatic Evaluation of Summaries",
    author = "Lin, Chin-Yew",
    booktitle = "Text Summarization Branches Out",
    month = jul,
    year = "2004",
    address = "Barcelona, Spain",
    publisher = "Association for Computational Linguistics",
    url = "https://aclanthology.org/W04-1013/",
    pages = "74--81"
}

@inproceedings{banerjee-lavie-2005-meteor,
    title = "{METEOR}: An Automatic Metric for {MT} Evaluation with Improved Correlation with Human Judgments",
    author = "Banerjee, Satanjeev  and
      Lavie, Alon",
    editor = "Goldstein, Jade  and
      Lavie, Alon  and
      Lin, Chin-Yew  and
      Voss, Clare",
    booktitle = "Proceedings of the {ACL} Workshop on Intrinsic and Extrinsic Evaluation Measures for Machine Translation and/or Summarization",
    month = jun,
    year = "2005",
    address = "Ann Arbor, Michigan",
    publisher = "Association for Computational Linguistics",
    url = "https://aclanthology.org/W05-0909/",
    pages = "65--72"
}

@inproceedings{popovic-2015-chrf,
    title = "chr{F}: character n-gram {F}-score for automatic {MT} evaluation",
    author = "Popovi{\'c}, Maja",
    editor = "Bojar, Ond{\v{r}}ej  and
      Chatterjee, Rajan  and
      Federmann, Christian  and
      Haddow, Barry  and
      Hokamp, Chris  and
      Huck, Matthias  and
      Logacheva, Varvara  and
      Pecina, Pavel",
    booktitle = "Proceedings of the Tenth Workshop on Statistical Machine Translation",
    month = sep,
    year = "2015",
    address = "Lisbon, Portugal",
    publisher = "Association for Computational Linguistics",
    url = "https://aclanthology.org/W15-3049/",
    doi = "10.18653/v1/W15-3049",
    pages = "392--395"
}

\clearpage
\appendix


\clearpage
\section{Review and Rebuttal Mapping Details}
\label{app:map}

\subsection{Example of Mapping}
\begin{figure*}[!t]
    \vspace*{-2em} 
    \centering
	\includegraphics[width=1.0\linewidth]{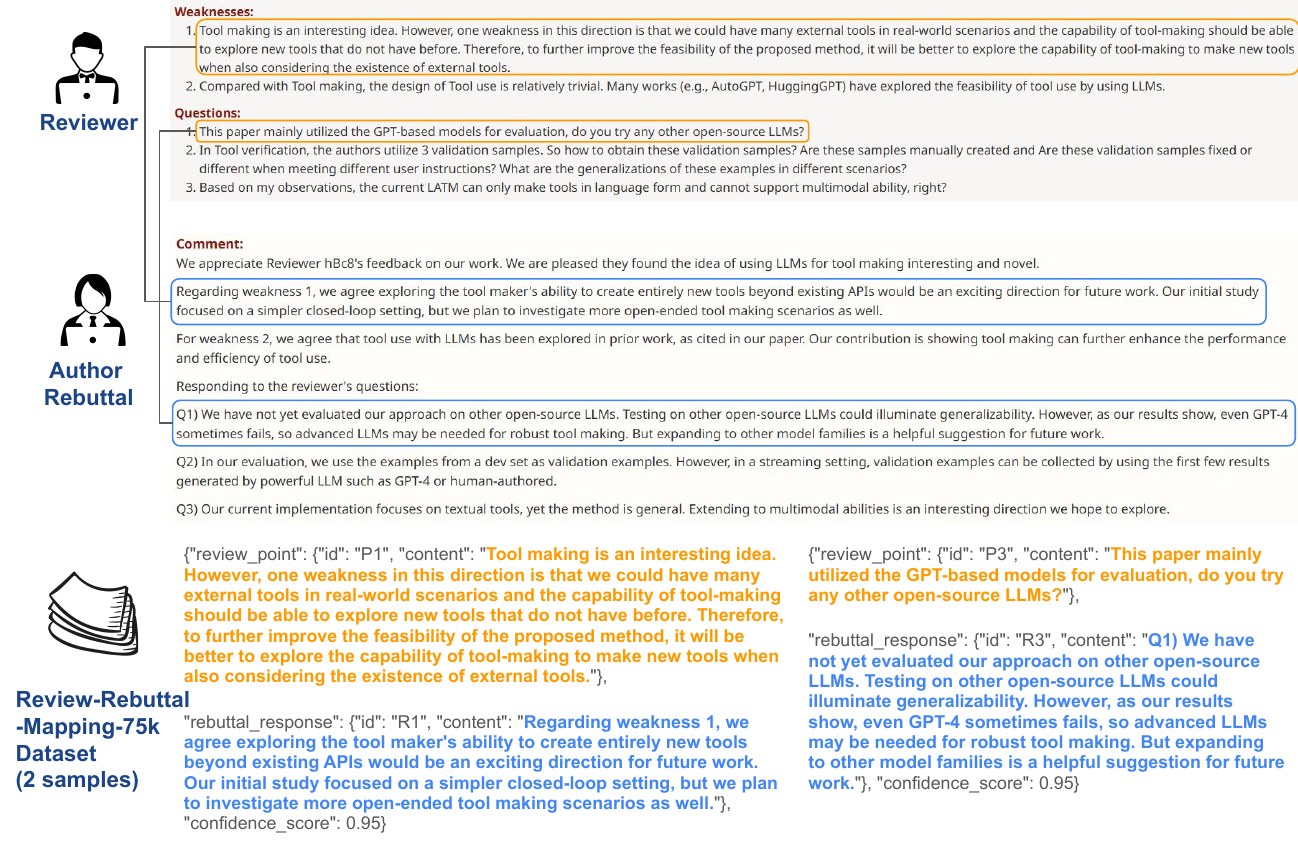}
	\caption{A mapping example of our \data\ dataset. The review and rebuttal are from the paper titled ``Large Language Models as Tool Makers''.}
	\label{fig:mappingexample}
\end{figure*}

Here is an example in our \data\ dataset in \autoref{fig:mappingexample}. We sample two mappings in the list.

\subsection{Prompts for Review Segment and Mapping}
The prompt for review segmentation is shown in \autoref{fig:promptseg}.
The prompt for mapping between review segment and rebuttal segment is shown in \autoref{fig:promptmap}.

\begin{figure*}[h]
\centering
\noindent
\begin{minipage}{\linewidth}
\begin{tcolorbox}[colback=black!7.5!white, colframe=black!30!white, fontupper=\footnotesize, fonttitle=\footnotesize]
\textcolor{blue}{\textbf{[System Prompt]}}

You are a professional academic review text analysis assistant. Your task is to segment a complete "Weaknesses \& Questions" section from an academic paper review into independent, specific points.\\

You must follow these rules:\\
1. Each point should be an independent, specific issue or weakness\\
2. Preserve the core meaning of the original text without adding or removing information\\
3. Maintain existing numbering structures if present (e.g., 1., 2., W1, W2, etc.)\\
4. Handle various formatting styles including:\\
   - Numbered lists (1., 2., 3.)\\
   - Letter prefixes (W1, W2, Q1, Q2)\\
   - Markdown bullet points (-, *, +)\\
   - Section headers (\#\# Weaknesses, \#\# Questions)\\
5. If no clear numbering exists, logically segment based on content structure\\
6. Each point should contain sufficient context to be understood independently\\
7. Preserve the original language and terminology used by the reviewer\\

\textcolor{blue}{\textbf{[User Prompt]}}

Please segment the following Weaknesses \& Questions text into independent points:\\

\texttt{\{weaknesses\&questions\_text\}} \\

IMPORTANT: Regardless of the input format (bullet points, numbered lists, paragraphs, etc.), you MUST output in this exact format:\\

Point 1: [Complete content of the first weakness point]\\
Point 2: [Complete content of the second weakness point]\\
Point 3: [Complete content of the third weakness point]\\
...\\

Rules:\\
- Use exactly "Point X:" where X is a number starting from 1\\
- Include ALL weakness points from the input, don't skip any\\
- Each point should be complete and independently understandable\\
- Preserve the original meaning and wording as much as possible\\
- If the input has bullet points (-, *, +) or numbered lists (1., 2.), convert them to Point format\\
- If the input has long paragraphs, break them into logical points\\
\end{tcolorbox}
\end{minipage}
\captionof{figure}{Prompt used to segment the weaknesses and questions parts of the review into segments.}
\label{fig:promptseg}
\end{figure*}

\begin{figure*}[h]
\centering
\noindent
\begin{minipage}{\linewidth}
\begin{tcolorbox}[colback=black!7.5!white, colframe=black!30!white, fontupper=\footnotesize, fonttitle=\footnotesize]
\textcolor{blue}{\textbf{[System Prompt]}}

You are a professional academic review analysis assistant. Your task is to perform precise one-to-one mapping between review weakness points and author rebuttal responses.\\

Guidelines for mapping:\\
1. Carefully analyze the rebuttal text to identify which sections respond to specific weaknesses\\
2. Look for explicit references (W1, W2, Point 1, etc.) or implicit topical connections\\
3. Extract the complete response content that addresses each weakness\\
4. Assign confidence scores (0-1) based on the clarity and directness of the mapping\\
5. Mark as "No Response" if a weakness is not addressed in the rebuttal\\
6. Be conservative with confidence scores - only use high scores (>0.8) when the mapping is very clear\\
7. Preserve the exact wording from the rebuttal when extracting responses\\

CRITICAL RULE - NO SHORTCUTS OR REFERENCES:\\
You must NEVER use summarizing phrases or references like "[Same content as W2 response]", "[Similar to above]", "[As mentioned earlier]", etc. Always copy the complete, verbatim text from the rebuttal for each weakness point, even if the same rebuttal section addresses multiple weaknesses. Repetition is required and expected - do not try to avoid it.\\

\textcolor{blue}{\textbf{[User Prompt]}}

Please map each weakness point in <weakness\_points> to its corresponding rebuttal response from <rebuttal\_text>:\\

<weakness\_points> \\
\texttt{\{weaknesses\_list\}} \\
</weakness\_points> \\

<rebuttal\_text> \\
\texttt{\{rebuttal\_text\}} \\
</rebuttal\_text> \\

MANDATORY REQUIREMENTS:\\
- Output a mapping line for EVERY weakness point (W1, W2, W3, ... up to W\{len(weakness\_points)\}) \\
- Use exactly "W[number] -> R[number]:" or "W[number] -> No Response" format\\
- Include confidence score in parentheses for every mapping\\
- Do not skip any weakness numbers\\
- If you cannot find a rebuttal response, write "No Response" instead of omitting the line\\

CRITICAL: You MUST provide a mapping for EVERY weakness point listed above. Do not skip any weakness points.\\
ABSOLUTELY CRITICAL - NO SUMMARIZATION OR SHORTCUTS FOR REBUTTAL CONTENT:\\
- ALWAYS copy the COMPLETE, FULL, VERBATIM text from the rebuttal for each weakness, even if content is identical to previous responses\\
- NEVER use summarizing or abbreviated phrases like "[Same content as W2 response]" or "[Similar to above]" or "[Full content identical to W5 response]" - always provide the complete original text\\
- Do NOT abbreviate, summarize, or reference other responses\\
- If the same rebuttal section addresses multiple weaknesses, copy the ENTIRE text in full for each relevant weakness\\
- NEVER add any commentary, explanation, or meta-text beyond the actual rebuttal content\\

<\/output\_format> (use EXACTLY this format):\\
W1 -> R1: [Specific rebuttal content addressing W1] (Confidence: 0.xx)\\
W2 -> R2: [Specific rebuttal content addressing W2] (Confidence: 0.xx)\\
W3 -> R3: [Specific rebuttal content addressing W3] (Confidence: 0.xx)\\
...continue for ALL weakness points...\\

If no rebuttal response exists for a weakness:\\
Wx -> No Response (Confidence: 1.0)\\
</output\_format>\\
\end{tcolorbox}
\end{minipage}
\captionof{figure}{Prompt used for mapping review segments with rebuttal segments.}
\label{fig:promptmap}
\end{figure*}

\section{Label Details}
\label{app:label}
We discuss details of our labeling of perspectives and impact categories here.

\subsection{Detailed Definition of Perspective labels and Impact Categories}
The detailed definition of perspective labels and rebuttal's impact categories are shown in~\autoref{tab:labels_detail}.
\begin{table*}[t]
\centering
\small
\setlength{\tabcolsep}{6pt}
\begin{tabularx}{\textwidth}{lX}
\toprule
\textbf{Label set} & \textbf{Definition} \\
\midrule
\multicolumn{2}{c}{\itshape Perspective of Review Segment} \\
\cmidrule(lr){1-2}
\emph{Experiments} & Concerns about experimental \textbf{setup and design}: missing or insufficient experiments, weak or absent ablations, unfair/weak baselines, unclear dataset descriptions or splits, problematic hyperparameter/seed choices, compute budgets, or training details that undermine empirical support. \\
\emph{Evaluation}  & How results are \textbf{measured, analyzed, and interpreted}: inappropriate/missing metrics, lack of statistical testing or error bars, insufficient analysis (e.g., no variance/sensitivity), unfair comparisons, or inconsistencies between reported numbers and claims. \\
\emph{Reproducibility} & Ability to \textbf{reproduce} results: missing implementation details or pseudo-code, absent code/data/links, unspecified hyperparameters, unclear preprocessing, seeds, or hardware; insufficient instructions to replicate tables/figures. \\
\emph{Novelty} & \textbf{Originality and relation to prior work}: overlap with existing methods, incremental contributions, unclear differentiation, missing or superficial positioning against closely related literature. \\
\emph{Theory} & \textbf{Theoretical correctness and justification}: flawed or unstated assumptions, gaps in proofs, incorrect derivations, mismatches between theorems and algorithms, or theory not supporting the claimed guarantees. \\
\emph{Writing} & \textbf{Clarity and readability}: grammar/style issues, ambiguous phrasing, undefined symbols/terms, confusing explanations, or organization that impedes understanding at the sentence/paragraph level. \\
\emph{Presentation} & \textbf{Figures/tables/organization}: unclear plots or legends, poor formatting, misplaced or redundant content, and overall paper structure that makes the narrative hard to follow. \\
\midrule
\multicolumn{2}{c}{\itshape Impact Level of Rebuttal Segment} \\
\cmidrule(lr){1-2}
\emph{CRP} & \textbf{Concrete Revision Performed}: authors point to specific changes or verifiable artifacts already added (new text/sections, new experiments/tables/figures, updated numbers, released code/data). \textit{Cues:} “We added/updated …”, “Section~X rewritten …”, “New ablation in Sec.~… shows …”, “Code/data at …”. \\
\emph{SRP} & \textbf{Specific Revision Plan}: authors commit to concrete future edits with where/what to change, but not yet implemented. \textit{Cues:} “We will add an ablation in Sec.~X …”, “We will redraw Fig.~…”, “We will clarify definitions in §…”. \\
\emph{VCR} & \textbf{Vague Commitment to Revise}: promises to improve without actionable details (no locations, artifacts, or timelines). \textit{Cues:} “We will revise accordingly.”, “We will improve clarity/writing.” \\
\emph{DWC} & \textbf{Defend Without Change}: argues the paper already addresses the point; no edits proposed. \textit{Cues:} “Already covered in Sec.~…”, “Setup is standard.”, “Claim stands.” \\
\emph{DRF} & \textbf{Deflect/Reframe}: shifts responsibility or reframes the issue; no change offered. \textit{Cues:} “Reviewer misinterprets …”, “Out of scope …”, “Reviewer phrasing is incorrect.” \\
\bottomrule
\end{tabularx}
\caption{Summary of \emph{perspective} labels for review segments and \emph{impact} labels for rebuttal segments in \S\ref{sec:taskdef}.}
\label{tab:labels_detail}
\end{table*}

\subsection{Perspective and Impact Category Distribution}
We analyze the seven-perspectives and impact category distribution in our RMR-75K dataset, which is shown in \autoref{tab:label_dist}.
\begin{table*}[!h]
\centering
\small
\setlength{\tabcolsep}{5pt}
\begin{tabular}{lccccc}
\toprule
\textbf{Perspective} & \textbf{CRP} & \textbf{SRP} & \textbf{VCR} & \textbf{DWC} & \textbf{DRF} \\
\midrule
Evaluation (11{,}257)      & 4{,}766 / 42.3\% &   903 / 8.0\% &  171 / 1.5\% & 5{,}249 / 46.6\% & 168 / 1.5\% \\
Experiments (25{,}160)     & 12{,}059 / 47.9\%& 2{,}272 / 9.0\%& 401 / 1.6\%  & 9{,}833 / 39.1\% & 595 / 2.4\% \\
Novelty (8{,}585)          & 2{,}828 / 32.9\% &   872 / 10.2\%& 185 / 2.2\%  & 4{,}578 / 53.3\% & 122 / 1.4\% \\
Presentation (4{,}776)     & 2{,}894 / 60.6\% &   803 / 16.8\%& 256 / 5.4\%  &   784 / 16.4\%   &  39 / 0.8\% \\
Reproducibility (4{,}402)  & 2{,}009 / 45.6\% &   465 / 10.6\%& 120 / 2.7\%  & 1{,}747 / 39.7\% &  61 / 1.4\% \\
Theory (12{,}822)          & 4{,}253 / 33.2\% & 1{,}110 / 8.7\%& 282 / 2.2\%  & 6{,}859 / 53.5\% & 318 / 2.5\% \\
Writing (8{,}540)          & 4{,}693 / 55.0\% & 1{,}149 / 13.5\%& 631 / 7.4\% & 1{,}997 / 23.4\% &  70 / 0.8\% \\
\midrule
\textbf{Overall}           & 33{,}502 / 44.3\%& 7{,}574 / 10.0\%& 2{,}046 / 2.7\%& 31{,}047 / 41.1\%& 1{,}373 / 1.8\% \\
\bottomrule
\end{tabular}
\caption{Label distribution by perspective (counts / \%).}
\label{tab:label_dist}
\end{table*}

\subsection{Prompt for labeling perspectives}
The prompt for labeling which perspective a review segment belongs to is shown in \autoref{fig:promptper}.
\begin{figure*}[t!]
\centering
\noindent
\begin{minipage}{\linewidth}
\begin{tcolorbox}[colback=black!7.5!white, colframe=black!30!white, fontupper=\footnotesize, fonttitle=\footnotesize]
\textcolor{blue}{\textbf{[System Prompt]}}

You are an expert in academic paper review classification. Your task is to identify from which perspective the reviewer is raising concerns or questions about the paper.\\

\textcolor{blue}{\textbf{[User Prompt]}}\\
The following review point is a weakness or question raised by a reviewer during peer review. Please classify this review point based on the PERSPECTIVE from which the reviewer is critiquing or questioning the paper:\\

1. **Experiments**: The reviewer is questioning experimental **setup and design**. This includes missing or insufficient experiments, lack of ablation studies, weak baseline comparisons, unclear descriptions of datasets, or issues with hyperparameter selection.\\

2. **Writing**: The reviewer is concerned about writing quality - grammar, clarity, readability, ambiguous phrasing, typos, missing definitions of symbols/terms, unclear explanations of concepts.\\

3. **Presentation**: The reviewer is critiquing presentation and organization - figures, tables, and organization issues, unclear plots, missing legends, poor formatting, misplaced content, overall paper structure making it hard to follow.\\

4. **Theory**: The reviewer is questioning theoretical aspects - incorrect mathematical derivations, flawed assumptions, weak theoretical justification, missing proofs, inconsistency between claims and formulas.\\

5. **Novelty**: The reviewer is questioning novelty and originality - lack of novelty or originality, overlap with prior work, incremental contribution, insufficient differentiation from existing methods.\\

6. **Reproducibility**: The reviewer is concerned about reproducibility - missing implementation details, absent code or pseudo-code, hyperparameters not specified, insufficient information to reproduce results.\\

7. **Evaluation**: The reviewer is concerned about how the experimental results are **measured, analyzed, and interpreted**. This includes the use of inappropriate or missing evaluation metrics, insufficient analysis of results, or inconsistencies between reported results and the paper's claims.\\

8.  **Miscellaneous**: Content that is not a direct review point (weaknesses, questions, suggestions) about the paper. This includes polite remarks, Summative or transitional comments, summaries of the paper's or review's content, or irrelevant text.\\

Please analyze the following review point and identify from which perspective the reviewer is raising their concern. Respond with ONLY the category name (Experiments, Writing, Presentation, Theory, Novelty, Reproducibility, Evaluation, Miscellaneous).\\

Review point to classify:\\
\texttt{\{weakness\_content\}}\\

Perspective:\\
\end{tcolorbox}
\end{minipage}
\captionof{figure}{Prompt used to label which perspective a review segment belongs to.}
\label{fig:promptper}
\end{figure*}

\subsection{Prompt for labeling impact categories}
The prompt for labeling which impact category a rebuttal segment belongs to is shown in \autoref{fig:promptimp}.
\begin{figure*}[t!]
\centering
\noindent
\begin{minipage}{\linewidth}
\begin{tcolorbox}[colback=black!7.5!white, colframe=black!30!white, fontupper=\footnotesize, fonttitle=\footnotesize]
\textcolor{blue}{\textbf{[System Prompt]}}

You are a precise, deterministic classifier for rebuttal responses.\\

\textcolor{blue}{\textbf{[User Prompt]}}\\
Return only a compact JSON object: \{"impact": "<one\_of:[CRP,SRP,VCR,DWC,DRF]>"\}.\\
Categories:\\
- CRP: Concrete revision already made or concrete, verifiable artifact provided (new text/sections, new experiments/tables/figures, code/data links, numbers).\\
  Cues: "We added/updated...", "Section X rewritten...", "New ablation in Sec. ... shows ...", "Code/data at ...".\\
- SRP: Specific revision plan committed, but not yet implemented; where/what to revise is specific.\\
  Cues: "We will add ablation in Sec. X...", "We will redraw Fig. ...", "We will clarify definitions in §...".\\
- VCR: Vague promise to revise; no concrete actions, locations, or artifacts.\\
  Cues: "We will revise accordingly.", "We will improve writing/clarity.".\\
- DWC: Defend current paper as-is; no new change proposed.\\
  Cues: "Already covered in Sec. ...", "Setup is standard", "Claim stands".\\
- DRF: Shift issue to reviewer or avoid underlying point; no change offered.\\
  Cues: "Reviewer misinterprets ...", "Out of scope ...", "Reviewer phrasing is incorrect".\\
  
\end{tcolorbox}
\end{minipage}
\captionof{figure}{Prompt used to label which impact category a rebuttal segment belongs to.}
\label{fig:promptimp}
\end{figure*}
\section{Additional Training Details}
\label{app:train_details}
All training details are shown in \autoref{tab:train-config-table}.
\paragraph{Framework and hardware.}
We train using \texttt{LLaMA-Factory} on NVIDIA H200 GPUs (141\,GB) with \texttt{bf16} compute. 
We apply LoRA adapters to attention and MLP projections (\texttt{\{q,k,v,o,gate,up,down\}}) on top of the Llama-3.1-8B-Instruct base model.
All runs use per-device batch size $1$ with gradient accumulation as specified below.

\paragraph{Memory and speed optimizations.}
Direct Preference Optimization (DPO) requires evaluating both the chosen and the rejected responses under the policy and a frozen reference model, which increases memory use compared with SFT. 
To enable a $32$k token context in DPO, we combine: (i) FlashAttention-2 for efficient attention; (ii) DeepSpeed ZeRO-2 with gradient checkpointing; and (iii) a 4-bit quantized frozen reference model. 
This configuration reduces activation and parameter memory, allowing full paper context while maintaining throughput.

\paragraph{Schedules.}
SFT trains for $3$ epochs ($ =4{,}989$ steps) with learning rate $1.0{\times}10^{-4}$ and a cosine scheduler (warmup ratio $0.10$). 
DPO initializes from the SFT checkpoint and trains for 2 epochs ($=2{,}728$ steps) with learning rate $1.0{\times}10^{-5}$ and a cosine scheduler (warmup ratio $0.05$). 
Wall-clock time on H200 is $\approx 120$ hours for SFT and $\approx 203$ hours for DPO.

\begin{table*}[h]
\centering
\small
\setlength{\tabcolsep}{6pt}
\begin{tabular}{lll}
\toprule
\textbf{Setting} & \textbf{SFT} & \textbf{DPO} \\
\midrule
Framework & LLaMA-Factory & LLaMA-Factory \\
Base model & Llama-3.1-8B-Instruct & Llama-3.1-8B-Instruct \\
Precision & bf16 & bf16 \\
GPU & H200 141\,GB & H200 141\,GB \\
Context limit (\texttt{cutoff\_len}) & 32{,}768 & 32{,}768 \\
Generation limit (\texttt{max\_new\_tokens}) & 512 & 512 \\
Dataset & \textsc{ReviewSeg-SFT-13k} & \textsc{ReviewPref-DPO-22k} \\
Objective & Token-level CE & DPO (sigmoid) \\
Epochs / Steps & 3 / $4{,}989$ & 2 / $2{,}728$ \\
LR / Scheduler & $1\mathrm{e}{-4}$ / cosine & $1\mathrm{e}{-5}$ / cosine \\
Warmup ratio & 0.10 & 0.05 \\
Batch (per-dev $\times$ accum) & $1 \times 8$ & $1 \times 16$ \\
Eff.\ batch & 8 & 16 \\
LoRA target & q,k,v,o,gate,up,down & q,k,v,o,gate,up,down \\
LoRA rank / $\alpha$ / dropout & 8 / 16 / 0.05 & 16 / 16 / 0.05 \\
Grad.\ checkpointing & Off & On \\
Reference model & N/A & Frozen SFT \\
Ref.\ quantization & N/A & 4-bit \\
DeepSpeed (ZeRO-2) & Optional & On \\
FA2 (FlashAttention-2) & On & On \\
Wall-clock time & $\approx 120$ h & $\approx 203$ h \\
\bottomrule
\end{tabular}
\caption{Key training configuration for SFT and DPO. ``Eff.\ batch'' is per-device batch size~$\times$ accumulation steps. 
DeepSpeed denotes ZeRO-2; FA2 denotes FlashAttention-2.}
\label{tab:train-config-table}
\end{table*}


\section{Evaluation Details}
\label{app:eval}

\subsection{Baseline Prompt}
\label{app:baseline}
\autoref{fig:baseline} shows the prompt for different LLM baselines generating review segments.

\begin{figure*}[h]
\centering
\noindent
\begin{minipage}{\linewidth}
\begin{tcolorbox}[colback=black!7.5!white, colframe=black!30!white, fontupper=\footnotesize, fonttitle=\footnotesize]
\textcolor{blue}{\textbf{[System Prompt]}}

You are a professional reviewer. Provide a comment such as weakness, question or suggestion on the given paper in 1 to 3 sentences.\\

\textcolor{blue}{\textbf{[User Prompt]}}\\
Request: From the perspective of xxx, provide a comment on the following paper.\\
\texttt{<PAPER\_CONTEXT>}\\

\end{tcolorbox}
\end{minipage}
\caption{Prompt used to generate review segments by different LLM baselines.}
\label{fig:baseline}
\end{figure*}

\subsection{Baseline Adaptation Details}
\label{app:baseline_adapt}

All baselines of ``Other Methods'' receive the same paper text and the same requested perspective, and are required to output \emph{exactly one} review segment following our unified format (Appendix~\ref{app:llmasjudgeprompt}).
When a baseline naturally produces longer outputs (e.g., multi-section reviews, multiple bullets, or agent traces), we apply a deterministic post-processing rule to extract one segment, described below.

\paragraph{MARG.}
MARG~\cite{marg} generates review feedback via multiple LLM instances (agents) that each read a portion of the paper and then aggregate intermediate discussion into final comments.
To adapt MARG to our single-segment, perspective-conditioned setting, we inject the requested perspective into the leader's instruction so that the final synthesized comment targets the requested perspective.
We output only the leader's final synthesized comment and discard intermediate agent messages.
If multiple bullets are produced in the final synthesis, we take the first bullet as the single segment.

\paragraph{LimGen.}
LimGen~\cite{limgen} focuses on generating limitations with suggestions for improvement.
We adapt LimGen by prompting it to generate limitations \emph{conditioned on the requested perspective}.
If the output contains multiple limitation items, we deterministically select the first item (including its paired suggestion when present) as the single segment.
We only normalize surface formatting (e.g., removing numbering or headings) and do not add new content.

\paragraph{DeepReviewer-14B.}
DeepReviewer-14B~\cite{deepreview} is trained to produce comprehensive, structured reviews.
To make it comparable in our setting, we prompt DeepReviewer-14B to produce a \emph{single} perspective-specific comment in our segment format.
If the model outputs a full multi-section review despite the instruction, we extract the subsection that best matches the requested perspective using a fixed heading-based mapping (e.g., \textit{Experiments/Empirical Evaluation} $\rightarrow$ experiments; \textit{Writing/Presentation/Clarity} $\rightarrow$ clarity; \textit{Impact/Significance/Novelty} $\rightarrow$ contribution/impact), and then take the first paragraph in that subsection as the segment.
We run DeepReviewer-14B in pure generation mode without external tools to keep inputs comparable.


\subsection{Human Expert Evaluation Protocol}
\label{app:humaneval}
This appendix describes the interface and rubric we used to evaluate review segments and actionable rebuttal guidance.
\subsubsection{Interface}
\begin{figure*}[!t]
    \vspace*{-2em} 
    \centering
	\includegraphics[width=1.0\linewidth]{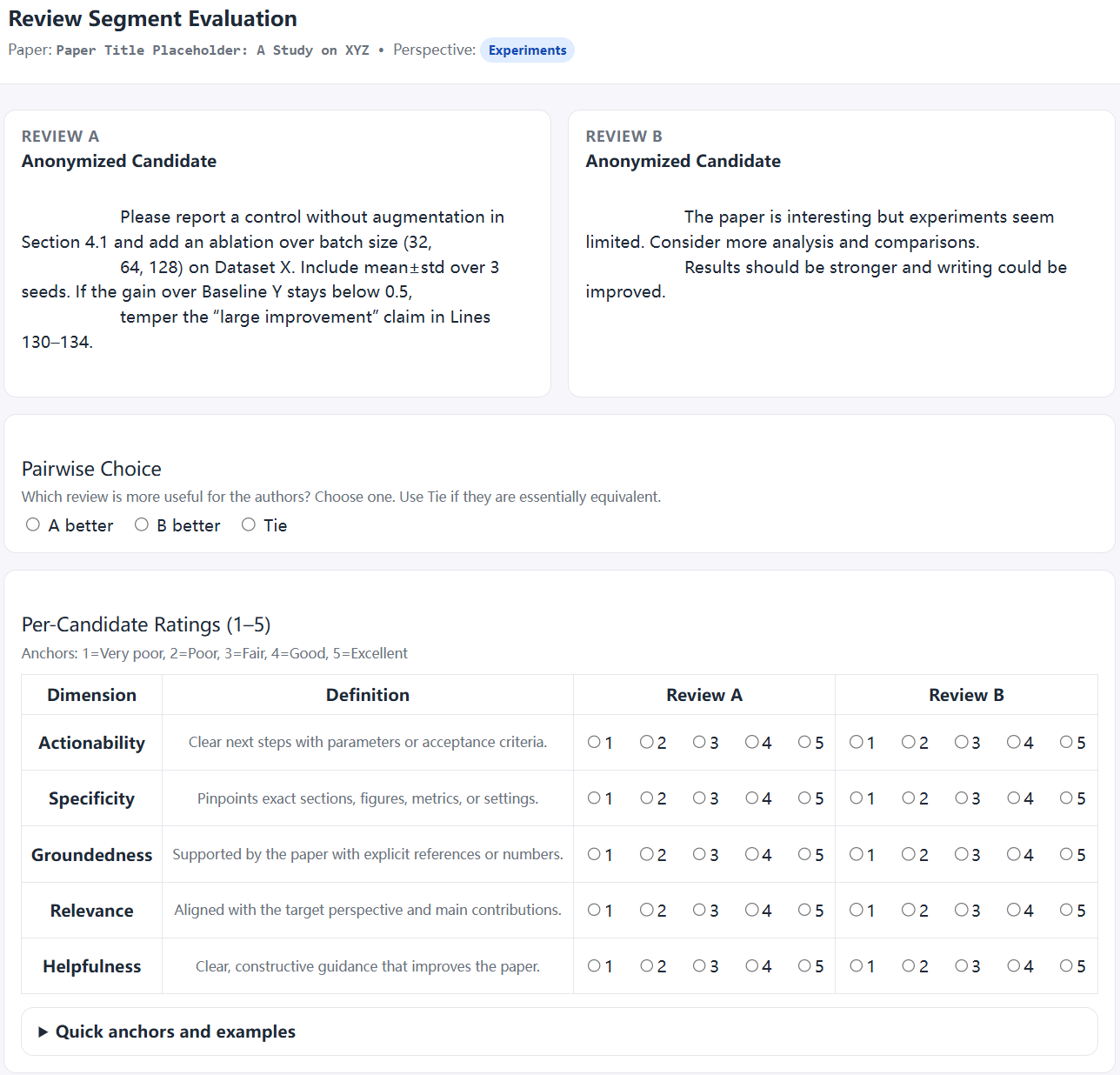}
	\caption{The interface of our human expert evaluation. The page contains 2 tasks: pairwise preference and per-candidate 1–5 ratings.}
	\label{fig:interface}
\end{figure*}

We collect two judgments per comparison on the page shown in \autoref{fig:interface}: (i) a \textit{pairwise preference}
between two anonymized candidate reviews (A vs.\ B vs.\ Tie), and (ii)
\textit{per-candidate 1--5 ratings} on five dimensions:
\textbf{Actionability}, \textbf{Specificity}, \textbf{Groundedness},
\textbf{Relevance}, and \textbf{Helpfulness}. Anchors are
$1=\text{Very poor}$, $2=\text{Poor}$, $3=\text{Fair}$, $4=\text{Good}$,
$5=\text{Excellent}$.\\
Raters judge only using evidence in the paper. They should avoid hallucinations and must not penalize a section for missing information if the same information exists elsewhere in the paper.

\subsubsection{Dimension Rubrics (Concise)}
\begin{enumerate}
    \item \textbf{Actionability}: Clear next steps with parameters or acceptance criteria as shown in \autoref{fig:actionability}.
    
    \begin{figure*}[!t]
        \vspace*{-2em} 
        \centering
    	\includegraphics[width=1.0\linewidth]{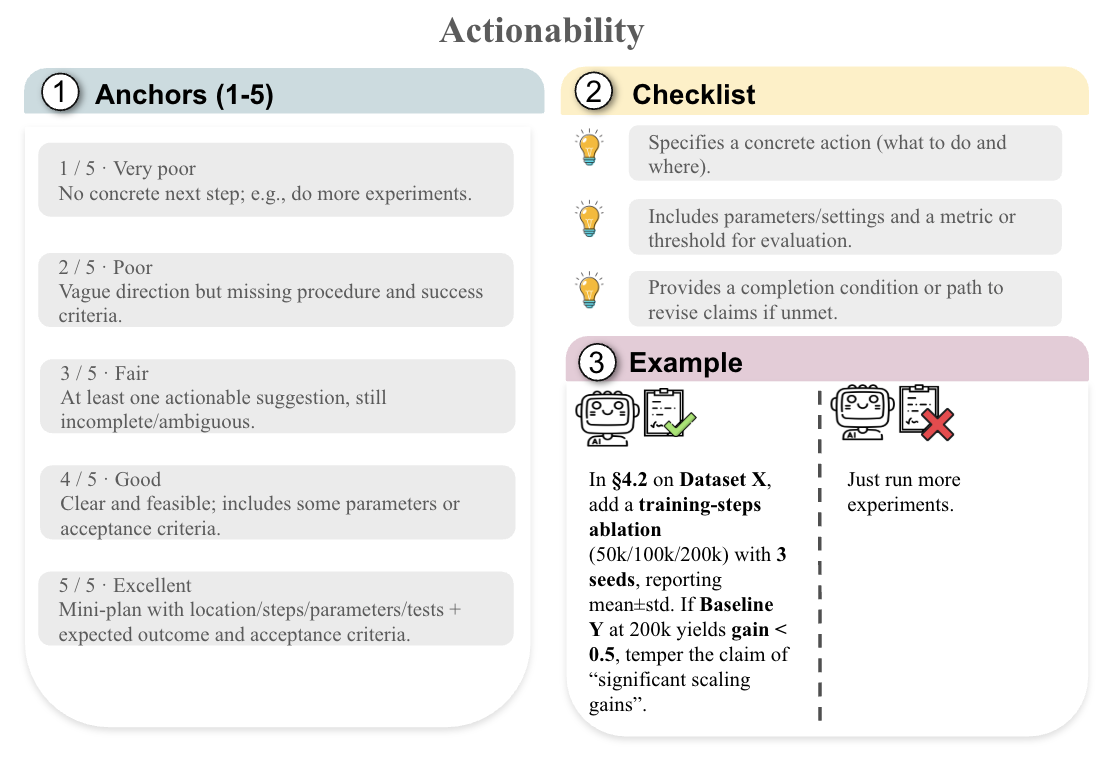}
    	\caption{Comparison guidelines for the “Actionability” criterion.}
    	\label{fig:actionability}
    \end{figure*}

    \item \textbf{Specificity}: Pinpoints exact sections, figures, metrics, or settings as shown in \autoref{fig:specificity}.
    
    \begin{figure*}[!t]
        \vspace*{-2em} 
        \centering
    	\includegraphics[width=1.0\linewidth]{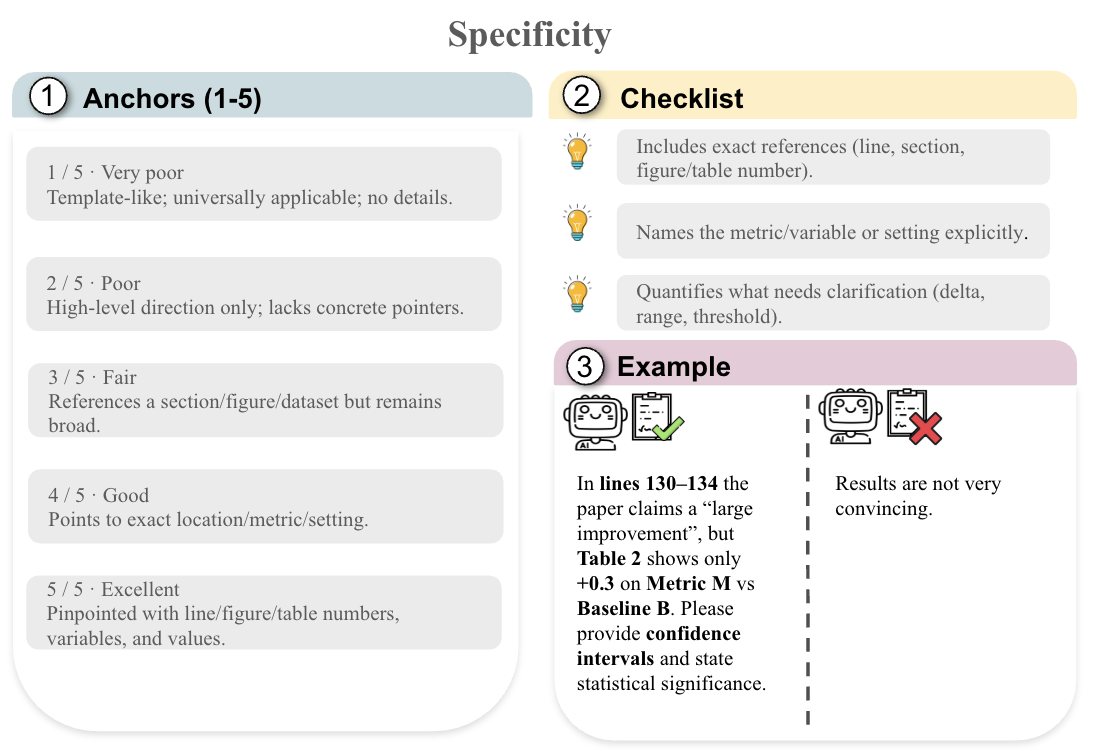}
    	\caption{Comparison guidelines for the “Specificity” criterion.}
    	\label{fig:specificity}
    \end{figure*}

    \item \textbf{Groundedness}:Supported by the paper with explicit references or numbers as shown in \autoref{fig:groundedness}.

    \begin{figure*}[!t]
        \vspace*{-2em} 
        \centering
    	\includegraphics[width=1.0\linewidth]{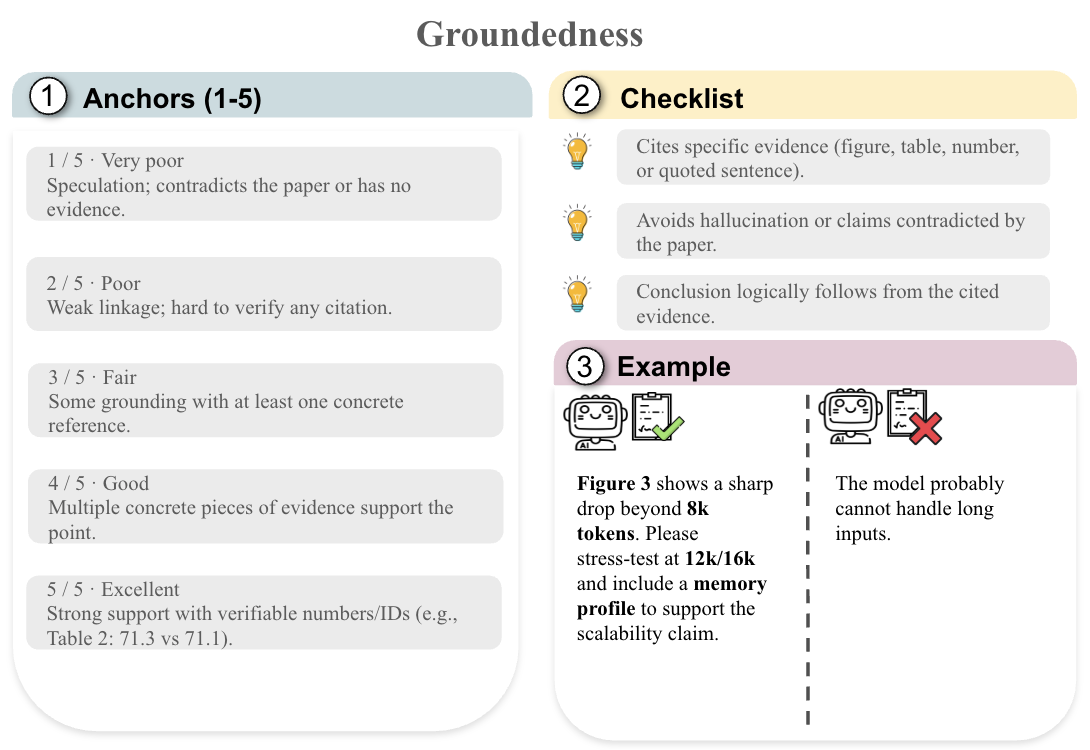}
    	\caption{Comparison guidelines for the “Groundedness” criterion.}
    	\label{fig:groundedness}
    \end{figure*}

    \item \textbf{Relevance}: Aligned with the target perspective and main contributions as shown in \autoref{fig:relevance}.

    \begin{figure*}[!t]
        \vspace*{-2em} 
        \centering
    	\includegraphics[width=1.0\linewidth]{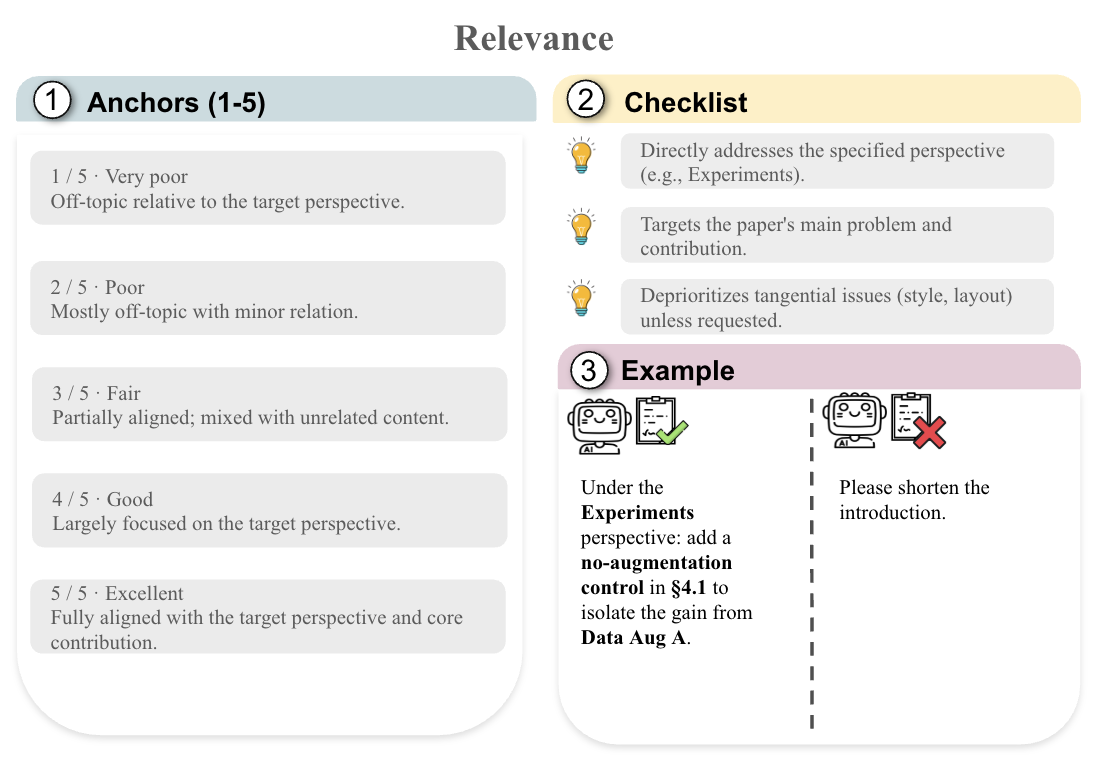}
    	\caption{Comparison guidelines for the “Relevance” criterion.}
    	\label{fig:relevance}
    \end{figure*}
    
    \item \textbf{Helpfulness}: Clear, constructive guidance that improves the paper as shown in \autoref{fig:helpfulness}.

    \begin{figure*}[!t]
        \vspace*{-2em} 
        \centering
    	\includegraphics[width=1.0\linewidth]{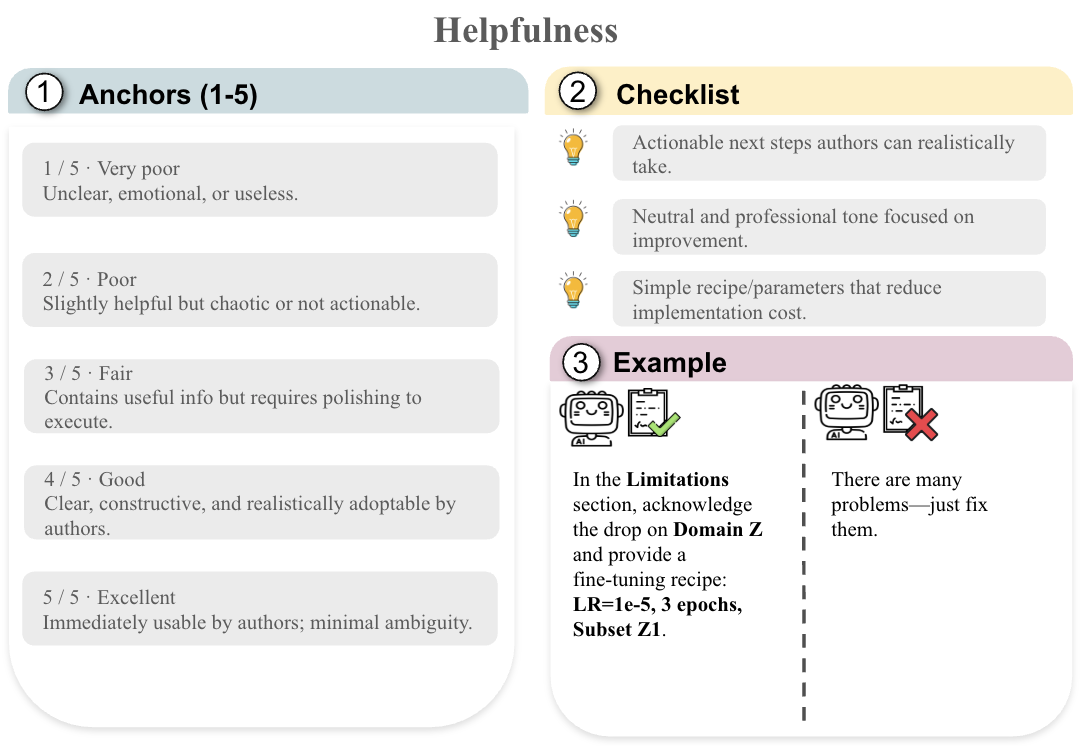}
    	\caption{Comparison guidelines for the “Helpfulness” criterion.}
    	\label{fig:helpfulness}
    \end{figure*}
\end{enumerate}

\subsection{LLM-as-a-Judge Evaluation}
\label{app:llmasjudgeprompt}
Here we discuss some details of LLM-as-a-judge evaluation.

The prompt for point-wise evaluation is \autoref{fig:promptllmpoint}.
\begin{figure*}[t!]
\centering
\noindent
\begin{minipage}{\linewidth}
\begin{tcolorbox}[colback=black!7.5!white, colframe=black!30!white, fontupper=\footnotesize, fonttitle=\footnotesize]
\textcolor{blue}{\textbf{[System Prompt]}}

You are an expert in evaluating peer review quality.\\
Your task is to assess a peer review comment from multiple dimensions and provide scores (1-5) with detailed reasoning for each dimension.\\

\textcolor{blue}{\textbf{[User Prompt]}}\\

Please evaluate the following peer review comment based on the scoring rubric provided.\\

Scoring Rubric\\

\#\#\# Actionability (1-5)\\
1. Very poor: No concrete next step. Vague remarks like "improve experiments."\\
2. Poor: A possible step is implied but not described. No criteria for success.\\
3. Fair: At least one concrete suggestion, but incomplete or underspecified.\\
4. Good: Clear, feasible steps with some parameters or success criteria.\\
5. Excellent: A short plan with steps, locations in the paper, parameters or tests, and what outcome would address the issue.\\

\#\#\# Specificity (1-5)\\
1. Very poor: Generic template text that could apply to any paper.\\
2. Poor: Mentions broad areas but no details.\\
3. Fair: Refers to a section, figure, dataset, or claim but stays broad.\\
4. Good: Points to exact sections, figures, metrics, or settings.\\
5. Excellent: Pinpoints precise passages or numbers and names exact variables, metrics, or ablation locations.\\

\#\#\# Groundedness (1-5)\\
1. Very poor: Speculative, incorrect, or contradicted by the paper.\\
2. Poor: Weak link to the paper; no verifiable reference.\\
3. Fair: Partly grounded with at least one reference to paper content.\\
4. Good: Well supported with references to specific content.\\
5. Excellent: Strongly supported with exact identifiers or numbers from the paper (for example "Table 2 shows 71.3 vs 71.1 and the claim of a large gain is not supported").\\

\#\#\# Relevance (1-5)\\
1. Very poor: Off topic relative to the target perspective or the main paper issues.\\
2. Poor: Mostly off topic with minor relevant content.\\
3. Fair: Partially aligned. Mixes relevant and irrelevant feedback.\\
4. Good: Mostly aligned with the target perspective.\\
5. Excellent: Fully aligned with the target perspective and the paper's main contributions.\\

\#\#\# Helpfulness (1-5)\\
1. Very poor: Unclear, hostile, or not useful.\\
2. Poor: Slightly useful but confusing or impractical.\\
3. Fair: Some useful content, needs refinement to be actionable.\\
4. Good: Clear, constructive, and practically useful.\\
5. Excellent: Directly helps the authors improve the paper with minimal ambiguity.\\

**Paper Content:**\\
\texttt{{paper\_content}}\\

**Review Perspective:** \\
\texttt{{perspective}}\\

**Review Comment to Evaluate:**\\
\texttt{{review\_text}}\\

Please provide scores (1-5) for each dimension along with your reasoning. Be critical and precise in your evaluation.\\

You MUST respond with a valid JSON object in the following format (no markdown code blocks, just raw JSON):\\
\{\{\\
    "actionability\_score": <1-5>,\\
    "actionability\_reasoning": "<brief explanation>",\\
    ......\\
\}\}\\

\end{tcolorbox}
\end{minipage}
\captionof{figure}{Prompt used for point-wise evaluation for LLM-as-a-judge.}
\label{fig:promptllmpoint}
\end{figure*}

The prompt for pairwise evaluation on Actionability is \autoref{fig:promptllmpair}.
\begin{figure*}[t!]
\centering
\noindent
\begin{minipage}{\linewidth}
\begin{tcolorbox}[colback=black!7.5!white, colframe=black!30!white, fontupper=\footnotesize, fonttitle=\footnotesize]
\textcolor{blue}{\textbf{[System Prompt]}}

You are an impartial judge comparing the actionability of two peer-review segments.
Actionability means the feedback gives concrete, specific, and feasible guidance that authors can directly implement. Prefer segments that:\\
-specify what to change (methods, experiments, analyses, writing),\\
-localize where to change (section/figure/table/scope),\\
-propose how to change (procedures, metrics, datasets, ablations, edits),\\
-include verifiable artifacts or acceptance criteria (e.g., code/data, new experiments, numbers to report).\\
Output JSON schema:\{
"winner": "A" | "B",
"justification": "1–2 sentences citing the most decisive actionable cues."
\}\\

\textcolor{blue}{\textbf{[User Prompt]}}\\
Task: Choose the more actionable review segment for the specified perspective. Remember: no ties.\\
Perspective:\\
\texttt{<PERSPECTIVE>}\\
Paper context:\\
\texttt{<PAPER\_CONTEXT>}\\
Segment A:\\
\texttt{<REVIEW\_SEGMENT\_A>}\\
Segment B:\\
\texttt{<REVIEW\_SEGMENT\_B>}\\
\end{tcolorbox}
\end{minipage}
\caption{Prompt used for pairwise evaluation for LLM-as-a-judge on Actionability.}
\label{fig:promptllmpair}
\end{figure*}

\begin{figure*}[t]
    \centering
    \includegraphics[width=\linewidth]
    {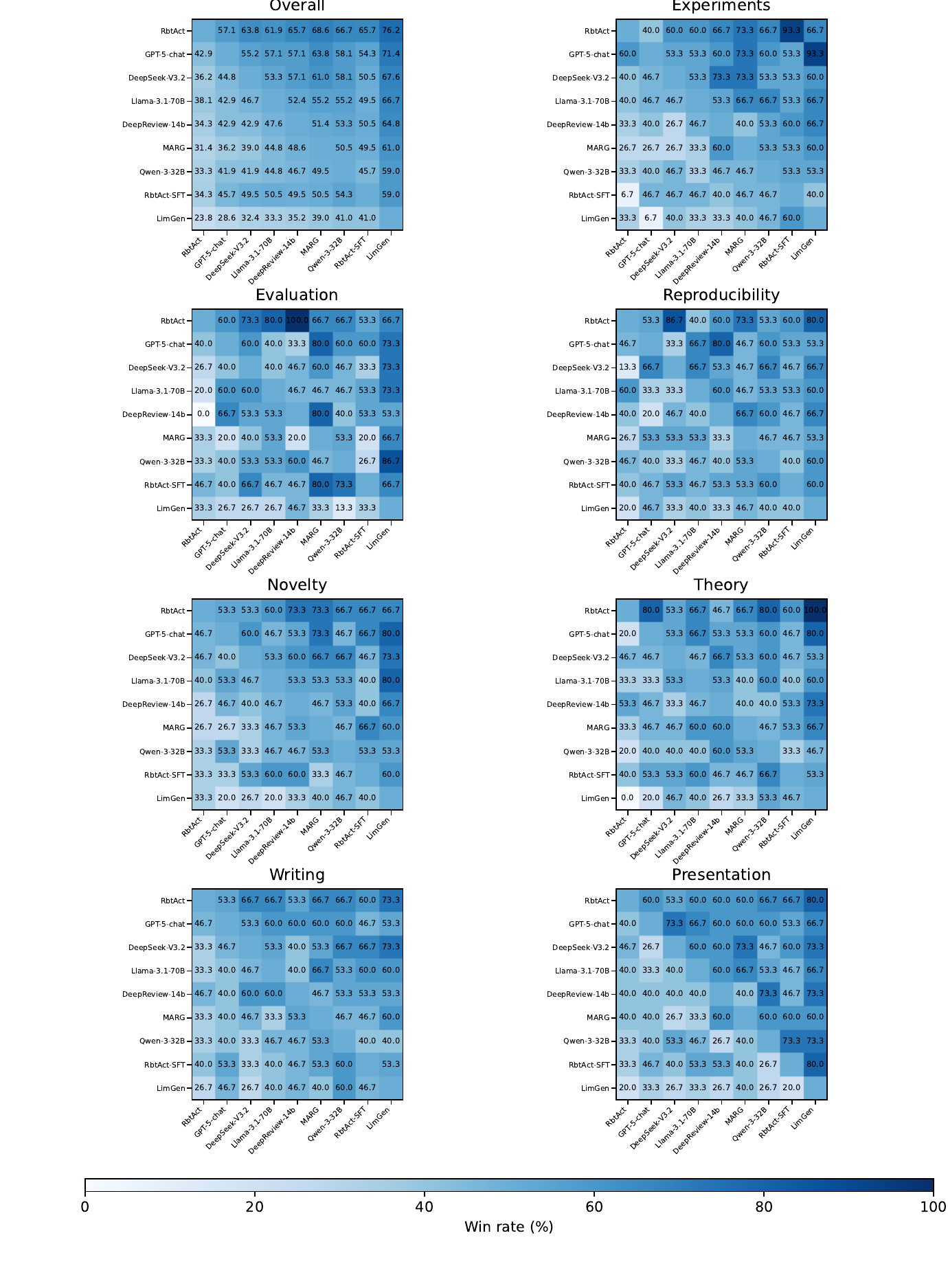}
    \caption{Pairwise win rates heatmaps by perspective (row beats column) of \S\ref{sec:expresults}}
    \label{fig:hm_overall}
\end{figure*}

To visualize pairwise evaluation results across seven different perspectives, \autoref{fig:hm_overall} shows heatmaps where each cell is the win rate of the \emph{row} model over the \emph{column} model.

\subsection{Automatic Evaluation}
\label{app:autoeval}
Besides human and LLM-as-a-Judge evaluation, we also evaluate our method and different baselines by some metrics. Results are shown in \autoref{tab:auto-metrics}.

As mentioned, we evaluate all 700 test instances from evaluation dataset with four metrics: BLEU@4 \cite{papineni-etal-2002-bleu}, ROUGE-L\textsubscript{sum} (F1) \cite{lin-2004-rouge}, METEOR \cite{banerjee-lavie-2005-meteor}, and chrF \cite{popovic-2015-chrf}.

Overall, our models perform competitively against baselines. \textsc{RbtAct} achieves the best ROUGE-L\textsubscript{sum} (12.64) and METEOR (11.65), while \textsc{RbtAct}-SFT attains the highest BLEU@4 (14.93). Both \textsc{RbtAct} variants outperform large proprietary and open models on these three metrics. On chrF, GPT-5-chat is strongest, and our models are close to each other (18.51 vs.\ 18.57).

Comparing \textsc{RbtAct} with \textsc{RbtAct}-SFT, differences are small. \textsc{RbtAct} slightly improves over SFT on ROUGE-L\textsubscript{sum} and METEOR. BLEU@4 is close to \textsc{RbtAct}-SFT. These trends indicate that rebuttal optimization refines phrasing but does not drastically change surface overlap. 

\begin{table}[h]
\centering
\small
\setlength{\tabcolsep}{3pt}
\renewcommand{\arraystretch}{1.1}
\begin{tabular}{lcccc}
\toprule
\textbf{Model} & \textbf{BLEU@4} & \textbf{ROUGE-L\textsubscript{sum}} & \textbf{METEOR} & \textbf{chrF} \\
\midrule
\textsc{RbtAct} (Ours) & 14.62 & \textbf{12.64} & \textbf{11.65} & 18.57 \\
\textsc{RbtAct}-SFT & \textbf{14.93} & 12.33 & 11.53 & 18.51 \\
\midrule
GPT-5-chat & 11.17 & 10.11 & 9.96 & \textbf{24.90} \\
DeepSeek-V3.2 & 10.19 & 9.76 & 8.49 & 17.78 \\
Llama-3.1-70B & 10.48 & 8.76 & 8.27 & 16.42 \\
Qwen-3-32B & 9.72 & 8.58 & 8.14 & 17.18 \\
\midrule
DeepReview & 12.40 & 11.35 & 10.21 & 19.69 \\
MARG & 10.95 & 9.12 & 8.64 & 18.16 \\
LimGen & 10.92 & 8.27 & 7.97 & 17.43 \\
\bottomrule
\end{tabular}%
\caption{Automatic evaluation on 700 test instances. Values are reported as percentages.}
\label{tab:auto-metrics}
\end{table}

\section{Case Study}
\label{app:casestudy}
Here we show some case studies to demonstrate why our model \textsc{RbtAct} is more actionable than other baselines in most cases from different perspectives. 
They are shown in \autoref{fig:casestudyall}.

\begin{figure*}[h]
\centering

\begin{blueBox}[{\footnotesize 
Case Study of Why \textsc{RbtAct} Reviews More Actionable
}]

\noindent
\begin{minipage}[t]{0.48\textwidth}
{\footnotesize
\textbf{\textsc{RbtAct} Review Feedback:}\\
The reported gains may be entangled with augmentation choices (MixUp/CutMix) and unquantified run-to-run variance, which leaves robustness in §5.2 not directly stress-tested; to disambiguate these factors, re-train without MixUp/CutMix under a fixed seed with three independent trials, report mean±std Top-1 in Table~3, and add a Corrupted ImageNet check aligned with the claims.
}
\end{minipage}\hfill
\begin{minipage}[t]{0.48\textwidth}
{\footnotesize
\textbf{Qwen-3-32b Review Feedback:}\\
The current training description under-specifies key regularization and sizing decisions, making it unclear how sensitive results are to setup and randomness. It would help to include a compact ablation that revisits weight decay and batch size.
}
\end{minipage}
\end{blueBox}

\begin{analysisbox}
\textbf{Analysis:}
It specifies exact factors to remove, run counts, where to report (Table 3), and a concrete robustness check—clear steps and success criteria.
\end{analysisbox}

\begin{blueBox}

\noindent
\begin{minipage}[t]{0.48\textwidth}
{\footnotesize
\textbf{\textsc{RbtAct} Review Feedback:}\\
In Figs.~2-3, small axis labels, palettes that are not color-blind-safe, missing y-axis units, and cramped legends reduce interpretability; revising the plots with adequately sized labels, an OKLCH-based color-blind-safe palette, explicit y-axis units, legends moved below the panels, and captions define each metric and sample size would make the figures verifiable.
}

\end{minipage}\hfill
\begin{minipage}[t]{0.48\textwidth}
{\footnotesize
\textbf{Llama-3.1-70b Review Feedback:}\\
Figure readability is limited by low apparent resolution, small typography, weak color separability, and inconsistent styling, with legend placement adding clutter. You may consider increasing DPI and font sizes, adopting more distinguishable colors under a unified style, clarifying captions in figure.~2.
}
\end{minipage}

\end{blueBox}

\begin{analysisbox}
\resizebox{\textwidth}{!}{
\textbf{Analysis:}
It names specific figures, thresholds, concrete palette guidance, and exact caption content—easy to implement and verify.
}
\end{analysisbox}

\begin{blueBox}

\noindent
\begin{minipage}[t]{0.48\textwidth}
{\footnotesize
\textbf{\textsc{RbtAct} Review Feedback:}\\
Extend Sec. 4.2: add baselines and SOTA with identical prompts; report macro-F1 and calibration with 95\% CIs via paired bootstrap over papers. Include error taxonomy in Appendix.
}

\end{minipage}\hfill
\begin{minipage}[t]{0.48\textwidth}
{\footnotesize
\textbf{MARG Review Feedback:}\\
Broaden evaluation to more datasets and compare against stronger models. Include confidence intervals where appropriate and a brief qualitative error analysis.
}
\end{minipage}

\end{blueBox}

\begin{analysisbox}
\resizebox{\textwidth}{!}{
\textbf{Analysis:}
It pins down section, comparators, metrics, CI method, and deliverables, giving precise implementation and acceptance checks.
}
\end{analysisbox}

\vspace{-6pt}
\caption{Case study comparing review feedback on Actionability from Experiment, Presentation and Evaluation perspectives.}
\label{fig:casestudyall}
\end{figure*}
\section{Additional Analyses}
\label{app:discuss}
\subsection{Severity and Paper-Strength Analyses.}
We further analyze whether rebuttal-derived supervision over-emphasizes minor issues and whether gains vary with paper strength.
We study (i) issue \emph{severity} using a perspective-based proxy and show that our training signal substantially covers major issues that are often defended, and (ii) the relationship between paper strength (OpenReview ratings) and actionability improvements.

\subsubsection{Are we biased toward minor issues? A severity proxy.}
We use a lightweight proxy for issue severity based on the review perspective labels.
We treat \textbf{major} issues as \{Experiments, Evaluation, Theory, Novelty, Reproducibility\} and \textbf{minor} issues as \{Writing, Presentation\}.
Table~\ref{tab:severity-impact} aggregates the impact-label counts by this proxy.

\begin{table}[t]
\centering
\small
\setlength{\tabcolsep}{4pt}
\renewcommand{\arraystretch}{1.05}
\begin{tabular}{lrr}
\toprule
Impact label & Major (Ev/Ex/Th/No/Re) & Minor (Wr/Pr) \\
\midrule
CRP & 25{,}915 (41.6\%) & 7{,}587 (57.0\%) \\
SRP & 5{,}622 (9.0\%) & 1{,}952 (14.7\%) \\
VCR & 1{,}159 (1.9\%) & 887 (6.7\%) \\
DWC & 28{,}266 (45.4\%) & 2{,}781 (20.9\%) \\
DRF & 1{,}264 (2.0\%) & 109 (0.8\%) \\
\bottomrule
\end{tabular}
\caption{Impact-label distribution by a perspective-based severity proxy.
Major issues are more frequently defended (DWC), but still exhibit substantial author uptake: CRP+SRP is 50.7\% for major issues and 71.6\% for minor issues.}
\label{tab:severity-impact}
\end{table}

As a result, major issues are indeed defended more often (DWC 45.4\% vs.\ 20.9\%), which matches the intuition that authors push back on higher-stakes critiques.
However, rebuttal supervision is \emph{not} dominated by minor issues: over half of major-issue mappings correspond to specific revisions (CRP+SRP = 50.7\%).
Moreover, our preference pairs are constructed \emph{within the same paper and perspective} and balanced across perspectives (\S\ref{sec:sft}), which reduces the risk that training is driven primarily by easy-to-fix minor edits.

\subsubsection{Do gains correlate with paper strength?}
\label{app:strength}

\paragraph{Setup.}
We operationalize \emph{paper strength} using the mean reviewer rating on OpenReview (averaged across reviewers for the same submission).
We split papers into three buckets (Weak / Medium / Strong) by tertiles of the mean rating.
On the 105-paper LLM-as-a-judge set, we compute the average Actionability score per bucket.

\begin{table}[t]
\centering
\small
\setlength{\tabcolsep}{2pt}
\renewcommand{\arraystretch}{1.05}
\begin{tabular}{lcccc}
\toprule
Bucket (by mean rating) & $n$ & \textsc{RbtAct} & Baseline & $\Delta$ (ours -- base) \\
\midrule
Weak (bottom 1/3)   & 35 & 3.35 & 3.08 & +0.27 \\
Medium (middle 1/3) & 35 & 3.33 & 3.11 & +0.22 \\
Strong (top 1/3)    & 35 & 3.46 & 3.27 & +0.19 \\
\bottomrule
\end{tabular}
\caption{Actionability by paper-strength bucket.}
\label{tab:strength-actionability}
\end{table}

\paragraph{Interpretation.}
The trend suggests that while stronger papers already elicit reasonably actionable feedback even from strong baselines, weaker papers benefit more from rebuttal-anchored supervision.
This is aligned with our training signal: rebuttals often make explicit which critiques lead to concrete revisions versus defenses, which helps the model prioritize actionable, fixable issues.

\subsection{Retrieval Baseline Analysis}
\label{app:retrieval-baseline}

To address the concern that the gains may come from a large organized corpus rather than learned generation, we add a simple retrieval baseline. Given a test paper and a target perspective, the baseline encodes the paper text as a query, searches only training review segments with the same perspective, excludes segments from the same paper to prevent leakage, and returns the nearest neighbor segment as the output. The baseline does not access rebuttals and uses the same single-segment output format as other systems.

Table~\ref{tab:retrieval-baseline} shows the LLM-as-a-judge results. Retrieval is competitive on relevance and specificity, which suggests that perspective-matched review segments provide useful topical priors. However, it performs worse than \textsc{RbtAct} on all five dimensions, especially Actionability, Groundedness, and Helpfulness. This indicates that nearest-neighbor reuse can retrieve plausible comments, but it often fails to adapt them to the paper's actual method, evidence, and experimental setting. In contrast, \textsc{RbtAct} learns to generate paper-specific and more actionable feedback from rebuttal-derived supervision.

\begin{table}[t]
\centering
\small
\setlength{\tabcolsep}{3.5pt}
\begin{tabular}{lccccc}
\toprule
\textbf{System} & \textbf{Action.} & \textbf{Spec.} & \textbf{Ground.} & \textbf{Rel.} & \textbf{Help.} \\
\midrule
\textsc{RbtAct} (ours) & 3.38 & 3.71 & 4.05 & 4.82 & 3.74 \\
Retrieval & 3.02 & 3.45 & 3.80 & 4.48 & 3.39 \\
\bottomrule
\end{tabular}
\caption{LLM-as-a-judge comparison with a retrieval baseline. The retrieval baseline returns the nearest training review segment under the same perspective while excluding segments from the same paper. Higher is better.}
\label{tab:retrieval-baseline}
\end{table}

\subsection{Ordinal-Aware Analysis of Human Ratings}
\label{app:ordinal-human}

Human pointwise ratings in~\autoref{tab:pointwise-combined} are collected on a 1--5 Likert scale. Since Likert ratings are ordinal, we provide an ordinal-aware analysis to complement the mean scores reported in the main paper. Specifically, we report the median and interquartile range, the percentage of ratings at least 4, and a ridit score computed from pooled category frequencies for Actionability and Helpfulness, as shown in~\autoref{tab:ordinal-summary}. The ridit score measures the relative standing of a system under the pooled ordinal distribution, where values above 0.5 indicate ratings that are stochastically higher than the pooled average.

\begin{table*}[h]
\centering
\small
\setlength{\tabcolsep}{4.5pt}
\renewcommand{\arraystretch}{1.08}
\begin{tabular}{lcccc@{\hspace{16pt}}cccc}
\toprule
& \multicolumn{4}{c}{\textbf{Actionability}} 
& \multicolumn{4}{c}{\textbf{Helpfulness}} \\
\cmidrule(r){2-5} \cmidrule(l){6-9}
\textbf{System} 
& \textbf{Mean} & \textbf{Med. [IQR]} & \textbf{\% $\geq$ 4} & \textbf{Ridit}
& \textbf{Mean} & \textbf{Med. [IQR]} & \textbf{\% $\geq$ 4} & \textbf{Ridit} \\
\midrule

\rowcolor{grpours}
\multicolumn{9}{c}{\textbf{Ours}} \\
\textsc{RbtAct} (ours) 
& \textbf{3.46} & \textbf{4 [2, 5]} & \textbf{52} & \textbf{0.548}
& 4.25 & 4 [2, 5] & 82 & 0.509 \\
\textsc{RbtAct}-SFT    
& 3.28 & 3 [2, 4] & 45 & 0.497
& 4.24 & 4 [1, 5] & 81 & 0.505 \\
\addlinespace[2pt]

\rowcolor{grpllm}
\multicolumn{9}{c}{\textbf{LLMs}} \\
GPT-5-chat      
& 3.38 & 3 [2, 5] & 47 & 0.523
& \textbf{4.49} & \textbf{5 [3, 5]} & \textbf{91} & \textbf{0.574} \\
DeepSeek-V3.2   
& 3.15 & 3 [2, 5] & 39 & 0.461
& 4.28 & 4 [2, 5] & 83 & 0.517 \\
Llama-3.1-70B   
& 3.22 & 3 [1, 4] & 42 & 0.478
& 4.15 & 4 [1, 5] & 77 & 0.482 \\
Qwen-3-32B      
& 3.06 & 3 [1, 4] & 35 & 0.437
& 4.12 & 4 [1, 5] & 75 & 0.472 \\
\addlinespace[2pt]

\rowcolor{grpother}
\multicolumn{9}{c}{\textbf{Other Methods}} \\
MARG            
& 3.20 & 3 [2, 5] & 41 & 0.474
& 4.18 & 4 [1, 5] & 79 & 0.491 \\
DeepReviewer-14B  
& 3.27 & 3 [2, 4] & 44 & 0.494
& 4.21 & 4 [1, 5] & 77 & 0.503 \\
LimGen          
& 3.16 & 3 [1, 4] & 40 & 0.465
& 4.05 & 4 [2, 5] & 73 & 0.450 \\
\bottomrule
\end{tabular}
\caption{Ordinal-aware summaries of human Likert ratings on Actionability and Helpfulness. Median, interquartile range, threshold rate, and ridit score are reported in addition to the descriptive mean. Higher is better.}
\label{tab:ordinal-summary}
\end{table*}

We further conduct paired nonparametric tests on per-item human ratings, as shown in~\autoref{tab:ordinal-tests}. For each of these two dimensions, Actionability and Helpfulness, ratings are compared between \textsc{RbtAct} and each baseline on the same evaluated instances using the Wilcoxon signed-rank test. We report the signed-rank statistic, two-sided $p$-value, and rank-biserial correlation as an effect size. Positive effects favor \textsc{RbtAct}. This analysis avoids assuming equal intervals between Likert categories.

\begin{table*}[h]
\centering
\small
\setlength{\tabcolsep}{6pt}
\renewcommand{\arraystretch}{1.08}
\begin{tabular}{lcccc@{\hspace{18pt}}cccc}
\toprule
& \multicolumn{4}{c}{\textbf{Actionability}} 
& \multicolumn{4}{c}{\textbf{Helpfulness}} \\
\cmidrule(r){2-5} \cmidrule(l){6-9}
\textbf{Comparison} 
& \textbf{$W$} & \textbf{$p$} & \textbf{Effect} & \textbf{Result}
& \textbf{$W$} & \textbf{$p$} & \textbf{Effect} & \textbf{Result} \\
\midrule
\textsc{RbtAct} vs. \textsc{RbtAct}-SFT
& 1812.5 & 0.018 & 0.23 & $+$ 
& 2148.0 & 0.641 & 0.04 & n.s. \\

\textsc{RbtAct} vs. GPT-5-chat
& 1987.0 & 0.276 & 0.09 & n.s.
& 1695.5 & 0.031 & -0.19 & $-$ \\

\textsc{RbtAct} vs. DeepSeek-V3.2
& 1654.0 & 0.004 & 0.31 & $+$
& 2186.5 & 0.784 & -0.02 & n.s. \\

\textsc{RbtAct} vs. Llama-3.1-70B
& 1875.5 & 0.031 & 0.20 & $+$
& 1922.0 & 0.096 & 0.14 & n.s. \\

\textsc{RbtAct} vs. Qwen-3-32B
& 1538.0 & 0.001 & 0.36 & $+$
& 1830.5 & 0.041 & 0.18 & $+$ \\

\textsc{RbtAct} vs. MARG
& 1819.5 & 0.022 & 0.22 & $+$
& 1994.0 & 0.184 & 0.11 & n.s. \\

\textsc{RbtAct} vs. DeepReviewer-14B
& 1936.0 & 0.086 & 0.15 & n.s.
& 2075.5 & 0.502 & 0.05 & n.s. \\

\textsc{RbtAct} vs. LimGen
& 1711.0 & 0.007 & 0.28 & $+$
& 1788.0 & 0.022 & 0.21 & $+$ \\
\bottomrule
\end{tabular}
\caption{Paired ordinal tests on human ratings. We use the Wilcoxon signed-rank test over matched evaluation instances and report rank-biserial correlation as the effect size. Positive effects favor \textsc{RbtAct}. $+$ indicates that \textsc{RbtAct} is significantly better, $-$ indicates that the baseline is significantly better, and n.s. indicates no significant difference.}
\label{tab:ordinal-tests}
\end{table*}

The ordinal-aware analysis is consistent with the main findings. \textsc{RbtAct} has the strongest actionability under the median-based summary and has positive paired effects against most baselines. Its improvement over \textsc{RbtAct}-SFT is supported by the Wilcoxon signed-rank test, while the helpfulness difference between the two systems is small and not significant. GPT-5-chat remains strongest on helpfulness, which is also reflected by its higher median, threshold rate, and ridit score. These results suggest that the main conclusions do not depend on treating Likert ratings as interval-scale measurements. 
\end{document}